\documentclass{article}

\usepackage{neurips_2025}

\usepackage[utf8]{inputenc} 
\usepackage[T1]{fontenc}    
\usepackage{hyperref}       
\usepackage{url}            
\usepackage{booktabs}       
\usepackage{amsfonts}       
\usepackage{nicefrac}       
\usepackage{microtype}      
\usepackage{xcolor}         
\usepackage{float}
\usepackage{wrapfig}
\usepackage{graphicx}
\usepackage{multirow}
\usepackage{pifont}
\usepackage{tocbasic}
\usepackage{titletoc}
\usepackage{enumitem}
\setlist[itemize]{left=5pt}

\nolinenumbers

\title{UniEval: Unified Holistic Evaluation for Unified Multimodal Understanding and Generation}

\author{%
  David S.~Hippocampus\thanks{Use footnote for providing further information
    about author (webpage, alternative address)---\emph{not} for acknowledging
    funding agencies.} \\
  Department of Computer Science\\
  Cranberry-Lemon University\\
  Pittsburgh, PA 15213 \\
  \texttt{hippo@cs.cranberry-lemon.edu} \\
}

\begin{document}

\maketitle

\vspace{-0.4cm}
\begin{abstract}

The emergence of unified multimodal understanding and generation models is rapidly attracting attention because of their ability to enhance instruction-following capabilities while minimizing model redundancy. However, there is a lack of a unified evaluation framework for these models, which would enable an elegant, simplified, and overall evaluation. Current models conduct evaluations on multiple task-specific benchmarks, but there are significant limitations, such as the lack of overall results, errors from extra evaluation models, reliance on extensive labeled images, benchmarks that lack diversity, and metrics with limited capacity for instruction-following evaluation. To tackle these challenges, we introduce UniEval, the first evaluation framework designed for unified multimodal models without extra models, images, or annotations. This facilitates a simplified and unified evaluation process. The \textbf{UniEval} framework contains a holistic benchmark, \textbf{UniBench} (supports both unified and visual generation models), along with the corresponding \textbf{UniScore} metric. UniBench includes 81 fine-grained tags contributing to high diversity. Experimental results indicate that UniBench is more challenging than existing benchmarks, and UniScore aligns closely with human evaluations, surpassing current metrics. Moreover, we extensively evaluated SoTA unified and visual generation models, uncovering new insights into UniEval’s unique values.
\end{abstract}

\vspace{-0.25cm}
\section{Introduction}
\vspace{-0.15cm}

\begin{figure}[htb]
  \centering
  \includegraphics[width=1.0\textwidth]{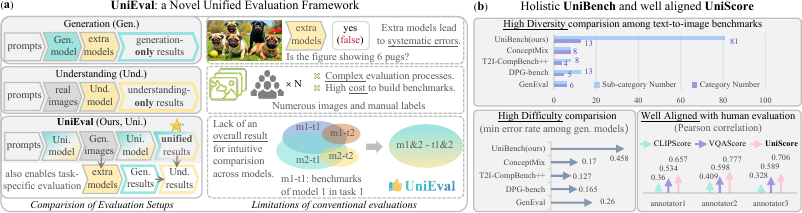}
  \vspace{-0.3cm}
  \caption{\label{fig_intro}\textbf{Overview of UniEval}. (\textbf{a}). The proposed UniEval unifies the evaluation of both the multimodal understanding and generation, eliminating limitations due to extra models, labeled images, and the lack of overall results. (\textbf{b}). The proposed UniBench is a holistic and challenging benchmark, with the UniScore metric aligning well with humans.}
  \vspace{-0.15cm}
\end{figure}

The unified multimodal understanding and generation models \cite{xie2024show,wu2024janus} are rapidly emerging. Many recent works \cite{chen2025janus,wu2024vila,ma2024janusflow,jiao2025unitoken} have proved that unified models can enhance instruction-following capabilities in visual generation \cite{ghosh2023geneval,hu2024ella} and reduce model redundancy. Although these models unify diverse tasks, their evaluations are still the same as task-specific models \cite{rombach2022high,betker2023improving,chen2023pixart,bai2025qwen2,chen2024internvl} relied on many conventional benchmarks \cite{ghosh2023geneval,hu2024ella,huang2025t2i,yue2024mmmu,li2023seed,liu2024mmbench,antol2015vqa,lu2022learn}. 

We cannot deny the value of task-specific evaluations via multiple benchmarks; however, current benchmarks \cite{hessel2021clipscore,ghosh2023geneval,xu2023imagereward,cho2024davidsonian,huang2025t2i,yue2024mmmu,liu2024mmbench,antol2015vqa} have significant limitations for unifed models: \textbf{1) Lack of overall results}: Averaging an overall result demands consistent benchmarks across models, which is usually not standardize in multitask settings, and many benchmarks raise the evaluation cost.
\textbf{2) Systemic error from extra models}: Automatic evaluations of visual generative \cite{hessel2021clipscore,ghosh2023geneval,xu2023imagereward,wu2024conceptmix} rely on extra models \cite{achiam2023gpt,hurst2024gpt,ye2023mplug}, and incorrect model predictions introduce unavoidable systemic errors (e.g., Wu et al \cite{wu2024conceptmix} reported error rate from 9.14\% to 25.8\% even for common attribute). \textbf{3) High resource costs}: Evaluations of multiple understanding benchmarks \cite{antol2015vqa,li2023seed,liu2024mmbench,yue2024mmmu} demand massive labeled images, where datasets in dozens of GB size make evaluation complex, and manual labeling makes building new datasets difficult. \textbf{4) Limited benchmark diversity and difficulty}: Existing visual generation datasets are limited in diversity and difficulty, restricting the evaluation of more advanced models (see Fig. \ref{fig_intro}b). \textbf{5) Insufficient metrics for instruction-following}: Conventional metrics like FID \cite{heusel2017gans}, IS \cite{salimans2016improved}, and CLIPScore \cite{hessel2021clipscore} are insufficient to evaluate instruction-following capabilities for complex prompts, which are crucial for unified models. So far, there is a lack of a unified benchmark for unified models, which would enable a simplified and overall evaluation.

To address these limitations, \emph{our motivation is to leverage the dual capabilities of unified models to evaluate themselves.} Specifically, the understanding part is applied to evaluate its visual generation without extra models, where their systematic errors are cleverly converted into Und. performance merged in the overall result. Meanwhile, generated images from the visual generation part eliminate massive labeled images, simplifying the evaluation process. This solution also yields an overall result, making model comparisons more intuitive and standardized. To support this motivation, an informative benchmark is needed to evaluate both understanding and generation. While current text-to-image benchmarks \cite{huang2025t2i,wu2024conceptmix,li2024evaluating,saharia2022photorealistic,cho2024davidsonian} just focus on some basic concepts, which are inadequate to evaluate understanding. Thus, we aim to build a holistic, challenging, and fine-grained compositional benchmark to evaluate unified models, emphasizing the key instruction-following capability.

\begin{figure}[htb]
  \centering
  \includegraphics[width=1\textwidth]{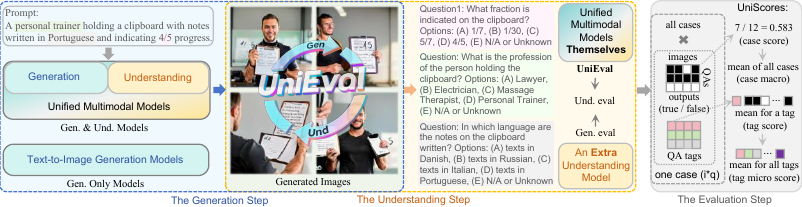}
  \vspace{-0.5cm}
  \caption{\label{fig_pipe}\textbf{Workflow of UniEval}. An example in UniBench processed by Janus-Pro-7B \cite{chen2025janus} to generate four images and outputs choices for each image and question (more examples in Appendix \ref{app_cases}). UniScores involves case-level accuracy in a case and tag-level accuracy from answers in the same tag. Our method is versatile, supporting generation evaluation with an extra model, and the understanding via the difference between unified and generation results (see Fig. \ref{fig_align_combine} and Appendix \ref{app_specific_eval}).}
  \vspace{-0.15cm}
\end{figure}

Based on the above motivation, we propose a unified evaluation framework called \textbf{UniEval} to evaluate unified multimodal models. It includes a holistic benchmark \textbf{UniBench} with the corresponding \textbf{UniScore} metric, which aligns well with human evaluations. We define 13 level-1 tags and 81 level-2 tags to ensure diversity, supporting evaluation in multiple aspects. Based on these tags, we extensively enumerate keywords, then generate 1,234 prompts and 4,231 question-answer (QA) pairs with manual quality control (see details in Fig. \ref{fig_bench}). In this benchmark, an evaluated model is required to generate four images for each prompt and output the correct options based on the given questions as shown in Fig. \ref{fig_pipe}. We calculate UniScores in both case-level and tag-level to analyze model performance, with the average of the level-1 tags as the overall UniScore. Our method is versatile, also allowing for task-specific evaluations through an extra model for generation evaluation. This enables a comparison between unified and generated results to analyze the specific understanding performance.

UniEval is the first unified evaluation framework and outperforms existing benchmarks in multiple aspects. Specifically, the 81 fine-grained tags of UniBench significantly exceed existing benchmarks \cite{cho2024davidsonian,hu2024ella} ($\leq$13). Moreover, UniBench presents higher difficulty, whose min error rate is 0.458, compared to 0.26 of GenEval and 0.165 of DPG-Bench (see Tab. \ref{tab_bench}). Notably, UniScore is an effective metric. Our multiple-choice metric correlates better with human evaluation on UniBench, achieving a Pearson correlation \cite{cohen2009pearson} of 0.716 with three annotators, compared to CLIPScore \cite{hessel2021clipscore} and VQAScore \cite{li2024evaluating}, which scored 0.372 and 0.575, respectively. Moreover, UniEval has about twice the model discriminability beyond task-specific benchmarks measured by the coefficient
of variation in Fig. \ref{fig_align_combine}, and it also supports separate analysis of Und. or Gen. From the leaderboards in Table \ref{tab_uni} and Table \ref{tab_gen}, UniEval reveals new insights in Table \ref{tab_insights}, highlighting its unique value. Our contributions include:

\begin{itemize}

\item We propose UniEval, the first evaluation framework for unified multimodal models, eliminating reliance on extra models and labeled images, achieving a simplified unified evaluation.

\item UniEval includes a holistic UniBench, which is currently the most challenging text-to-image benchmark with the highest number of fine-grained tags. The corresponding UniScore metric aligns well with human evaluations, surpassing existing metrics.

\item We conducted extensive evaluations on SoTA unified models and visual generation models, highlighting that UniEval can provide new insights with its unique values.
\end{itemize}

\begin{table}[htbp]
  \vspace{-0.15cm}
  \caption{\label{tab_bench}\textbf{Benchmark Comparision}. UniBench offers the most extensive tags and sub-tags in compositional text-to-image generation benchmarks, achieving high diversity. UniBench provides five related choices to minimize random error beyond binary options. UniBench has high difficulty, leading to a higher error rate of the SoTA model and more room for improvement. UniBench includes new features like generation evaluation, image-free, and annotation-free beyond Und. benchmarks.}
  \setlength{\tabcolsep}{2pt}
  \label{tab_bench_compare}
  \scriptsize
  \centering
  \begin{tabular}{ccccccc|cccc}
    \hline
    \textbf{Compositional} & \multicolumn{3}{c}{\textbf{Diversity}} & \multicolumn{2}{c}{\textbf{Difficulty}} & \multirow{2}{*}{\textbf{Avg. Rank}} & \textbf{Representative} & \multicolumn{3}{c}{\textbf{New Features}} \\
    \textbf{Gen. Benchmark} & Tags & Sub-Tags & Prompts & Options. & SoTA Error &  & \textbf{Und. Benchmark} & Gen. Eval & Img-Free & Anno-Free \\
    \hline
    T2I-CompBench++ \cite{huang2025t2i} & 4 & 8 & \textbf{8,000} & 2 & 0.127 & 4 & VQA \cite{antol2015vqa} & \ding{55} & \ding{55} & \ding{55} \\
    GenAI-Bench \cite{li2024evaluating} & 8 & 8 & 1,600 & 2 & 0.29 & 2.4 & GQA \cite{hudson2019gqa} & \ding{55} & \ding{55} & \ding{55} \\
    DSG-1K \cite{cho2024davidsonian} & 4 & 13 & 1,060 & 2 & 0.161 & 4 & SEED \cite{li2023seed} & \ding{55} & \ding{55} & \ding{55} \\
    ConceptMix \cite{wu2024conceptmix} & 8 & 8 & 300 & 2 & 0.17 & 3.8 & MMBench \cite{liu2024mmbench} & \ding{55} & \ding{55} & \ding{55} \\
    GenEval \cite{ghosh2023geneval} & 6 & 6 & 553 & 2 & 0.26 & 4.4 & ScienceQA \cite{lu2022learn} & \ding{55} & \ding{55} & \ding{55} \\
    DPG-Bench \cite{hu2024ella} & 5 & 13 & 1,000 & 2 & 0.165 & 3.8 & MMMU \cite{yue2024mmmu} & \ding{55} & \ding{55} & \ding{55}\\
    \textbf{UniBench (ours)} & \textbf{13} & \textbf{81} & 1,234 & \textbf{5} & \textbf{0.458} & \textbf{1.4} & \textbf{UniBench (ours)} & \ding{52} & \ding{52} & \ding{52} \\
    \hline
  \end{tabular}
  \vspace{-0.4cm}
\end{table}

\section{Related Works}
\vspace{-0.15cm}
\noindent\textbf{Unified Multimodal Understanding and Generation.} Conventional generative models typically generate texts for understanding tasks via multimodal LLMs \cite{liu2023visual,achiam2023gpt,bai2025qwen2,chen2024internvl}, or generating images via diffusion models \cite{rombach2022high,podell2023sdxl,betker2023improving,chen2023pixart,flux2024}. Some works \cite{dong2023dreamllm,ge2023planting,ge2023making,ge2024seed,ye2024x} equip multimodal LLMs with pre-trained diffusion models for unified generation. More recent methods try to train end-to-end unified multimodal models \cite{wu2024vila,jiao2025unitoken,wu2024janus,chen2025janus,xie2024show,xu2025show,li2024dual,wang2024illume,qu2024tokenflow,zhuang2025vargpt,zhou2024transfusion} to reduce model redundancy and enhance instruction-following capabilities in visual generation. These unified models are rapidly emerging and attract much attention, such as DeepSeek Janus-Pro \cite{chen2025janus} and OpenAI GPT-4o \cite{gpt-4o} (native image generation, API not available yet). In this paper, we conducted extensive evaluations for both unified (Table \ref{tab_uni}) and generation methods (Table \ref{tab_gen}), except for some unavailable models.

\noindent\textbf{Benchmarks.} 
Evaluation of unified models typically involves multiple benchmarks for each tasks. For example, using benchmarks like ScienceQA \cite{lu2022learn}, MMMU \cite{yue2024mmmu}, etc \cite{antol2015vqa,hudson2019gqa,chen2015microsoft,shah2019kvqa,mishra2019ocr,mathew2021docvqa,li2023seed,liu2024mmbench} to assess understanding capabilities, which rely on numerous images and labels. Our UniBench eliminates these dependencies, as a novel VQA benchmark for generated images. For the evaluation of generation models, image quality \cite{heusel2017gans,barratt2018note,xu2023imagereward,tian2025quality} assessments on general image benchmarks \cite{deng2009imagenet,lin2014microsoft,schuhmann2022laion} are widely used with other factors like alignment \cite{hessel2021clipscore}, fairness \cite{lee2023holistic}, style \cite{peng2024dreambench++}, etc \cite{bakr2023hrs}. Differently, unified models focus on instruction-following capabilities, making benchmarks like GenEval \cite{ghosh2023geneval}, DPG-Bench \cite{hu2024ella}, T2I-CompBench++ \cite{huang2025t2i}, and other text-to-image evaluations \cite{bakr2023hrs,wu2024conceptmix,higgins2017scan,li2024evaluating,saharia2022photorealistic,fengtraining,cho2024davidsonian} particularly relevant, with considered attributes like object, counting, colors, position, etc. Compared to these benchmarks, our UniBench evaluates many more aspects (see Table \ref{tab_bench}) to enhance the diversity, with greater difficulty and improvement potential. Most importantly, this is the first unified and elegant benchmark to evaluate both understanding and generation.

\noindent\textbf{Evaluation Metrics.} The accuracy for VQA base benchmarks \cite{yue2024mmmu,antol2015vqa,li2024evaluating,ghosh2023geneval} is the most common metric to evaluate understanding models, with some NLG metrics \cite{papineni2002bleu,denkowski2011meteor} for text generation tasks. Image generation quality assesment metrics like signal-to-noise ratio, FID \cite{heusel2017gans}, IS \cite{barratt2018note}, ImageReward \cite{xu2023imagereward} are widely used in generation-only models \cite{rombach2022high,podell2023sdxl,betker2023improving,chen2023pixart}. While unified models \cite{chen2025janus,xu2025show,wu2024vila} focus more on the instruction-following capacity using the CLIPScore \cite{hessel2021clipscore} or accuracy-based scores (e.g., VQAScore \cite{ghosh2023geneval}) on compositional text-to-image (T2I) benchmarks \cite{ghosh2023geneval,huang2025t2i,hu2024ella,cho2024davidsonian,wu2024conceptmix}. Our UniScore is also an accuracy-based base score, while we provide multiple choices rather than a binary choice about keyword existence. This difference reduces the random error, and Fig. \ref{fig_human_eval} suggests UniScore align well with human evaluations beyond other T2I metrics.

\vspace{-0.15cm}
\section{UniEval}
\vspace{-0.15cm}

UniEval is a novel unified evaluation framework featuring the holistic UniBench and the associated UniScore metrics, as shown in Fig. \ref{fig_pipe}. We elaborate on this framework by detailing UniBench in Sec. \ref{sec_31} and describing the UniScore metrics in Sec. \ref{sec_32}. Finally, we present a human study in Sec. \ref{sec_33}, suggesting UniScore aligns well with human perception beyond other metrics.
\vspace{-0.15cm}

\subsection{UniBench}
\label{sec_31}
\vspace{-0.15cm}


To achieve a unified evaluation, we need to construct a sufficiently diverse dataset that not only evaluates generation but also reflects various aspects of understanding ability. However, existing compositional benchmarks \cite{ghosh2023geneval,li2024evaluating,huang2025t2i,wu2024conceptmix} lack sufficient diversity; several attributes, such as color, objects, number, and position, are not enough to reflect a model’s understanding capabilities. Therefore, we have constructed a more holistic benchmark called UniBench via four steps.

\begin{figure}[htb]
  \centering
  \vspace{-0.15cm}
  \includegraphics[width=1\textwidth]{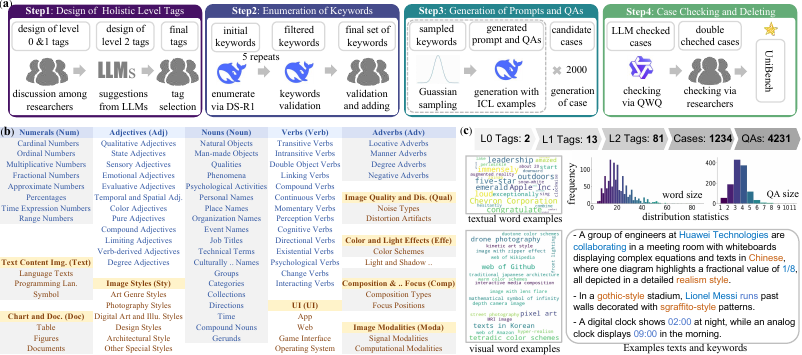}
  \vspace{-0.4cm}
  \caption{\label{fig_bench}\textbf{UniBench.} (\textbf{a}). It is built by researchers with LLMs in four steps (details in Appendix \ref{app_unibench_gen}). (\textbf{b}). We designed holistic level-1 tags and level-2 tags with many novel attributes. (\textbf{c}). Details of UniBench, including data size, distribution of words and QAs, examples of keywords and prompts (see more keywords in Appendix \ref{app_keywords} and prompts in Appendix \ref{app_unibench_gen}).}
  \vspace{-0.15cm}
\end{figure}

\textbf{Step 1}. Firstly, from the textual aspect, we selected five parts of speech suitable for image generation as level-1 tags. These include numerals, adjectives, nouns, verbs, and adverbs. From the visual perspective, we defined eight level-1 tags, including text content, chart and documents, image styles, image quality and distortion, color and lighting effects, composition and visual focus, image modality, and UI. Through collaboration between researchers and multiple large language models (LLMs), we further developed 81 hierarchical level-2 tags based on these 13 level-1 tags, as shown in Fig. \ref{fig_bench}b. These tags not only cover the attributes present in existing T2I benchmarks \cite{ghosh2023geneval,li2024evaluating,huang2025t2i,wu2024conceptmix} but also introduce many novel attributes, such as \emph{time, emotions, celebrities, events, locations, actions, degrees, languages, symbols, programming, modalities, charts, figures, documents, UI, noise types, color schemes, lighting effects, composition, visual focus, and other new attributes}. Additionally, UniBench includes highly challenging \emph{reasoning} tasks, like culturally specific names, which require reasoning about the region and personal appearances from names; and tasks requiring professional \emph{knowledge} such as programming languages, operating systems, computational modalities, etc.

\textbf{Step 2}. We enumerated extensive keywords for each level-2 tag via Deepseek-R1-70B \cite{guo2025deepseek}. The enumeration is repeated 5 times, and unique keywords were dropped to ensure reliability. Then we used this LLM to select keywords that are drawable as Appendix \ref{app_unibench_gen}. Finally, we checked keywords manually with added ones to finalize the keyword set (n = 3,285). We show some keywords as shown in Fig. \ref{fig_bench}c and more example keywords of each level-2 tag in Appendix \ref{app_keywords}.

\textbf{Step 3}. We bind keywords into prompts with corresponding questions and options in step 3. First, we randomly sample n level-2 tags using Gaussian sampling (std \& mean = 6), along with four keywords in this tag as options. Then, we use Deepseek-R1-70B to select a suitable keyword from each option pair to generate suitable prompts with a corresponding question and given options. In this process, we set prompt generation and question generation as two separate tasks, providing detailed task requirements, background, criteria, and several in-context examples. These requirements include ensuring the logical coherence of the prompts, ensuring that the keywords can be conveyed in the image, avoiding irrelevant content, and ensuring that questions can only be inferred from the image, etc. Please refer to more details and prompts to Appendix \ref{app_unibench_gen}.

\textbf{Step 4}. We filtered the 2,000 cases generated in Step 3 to ensure the quality of prompts and QAs (Q\&A). First, we used another LLM, QWQ-32B \cite{qwq32b}, for validation involving prompt verification and QA verification. Each step included detailed backgrounds, requirements, criteria, and in-context examples. We instructed the model to ensure prompts are drawable without being overly complex. For questions, we required strict validation to ensure options were directly derived from the image and related to keywords. Detailed prompts are given in Appendix \ref{app_unibench_gen}. Ultimately, we manually validated to finalize 1,234 prompts and 4,231 QAs (see examples in Fig. \ref{fig_bench}c and Appendix \ref{app_unibench_gen}).

\vspace{-0.15cm}
\subsection{UniScore}
\vspace{-0.15cm}
\label{sec_32}


This section focuses on the data flow of UniEval to introduce how we calculate the UniScore metrics. As shown in Fig. \ref{fig_pipe}, we first use the unified model's text-to-image capability to generate $i$ images ($i=4$) to reduce random error. Next, we utilize the model's image-to-text understanding to answer $q$ questions for each image, resulting in $i*q$ predictions. For each question, four options (A-D) belong to the same level-2 tag as the keyword, with an ``N/A or Unknown'' option for failed keyword generation. These five options help reduce random errors of binary options in other benchmarks \cite{wu2024conceptmix,cho2024davidsonian,hu2024ella,ghosh2023geneval}. To enhance UniEval's applicability, we support evaluating visual generation models with an extra Und. model in the same data flow as the unified model. Then, we analyze the understanding ability by the difference between unified and generation results, detailed in Appendix \ref{app_specific_eval}.

After obtaining outputs of one case (true or false), we can calculate the accuracy of a case as a case-level UniScore. By averaging this across all 1,234 cases, we obtain the case macro UniScore. Additionally, we provide more case-level UniScores categorized by the number of words and QAs for analyzing the differences among few, middle, and many (middle words [15-23], middle QAs [3-4]). Once all questions ($n=4,231$) have been processed, we get all outputs as $\textbf{o}$ ($n=4*4,231$). Then we aggregate outputs belonging to the same level-2 tag as $\textbf{o}$ and calculate its average accuracy to obtain the level-2 tag UniScore. By averaging the scores of level-2 tags under level-1, we derive the level-1 tag micro UniScore. Table \ref{tab_uni} and Table \ref{tab_gen} report the tag-level UniScores for all 13 level-1 tags, as well as their average as the final overall UniScore. We formulate the final UniScore as follows:
\vspace{-0.1cm}
\begin{equation}
s = \frac{1}{n}\sum^n_{i=1}\textbf{s}^1_i, \ \textbf{s}^1_i = \frac{1}{m}\sum^m_{j=1}\textbf{s}^2_j, \ \textbf{s}^2_j = \frac{1}{k}\sum^k_{l=1}\textbf{o}_l,
\end{equation}
\vspace{-0.1cm}
where $s$ is the final UniScore, averaged from $n$ level-1 tag UniScore $\textbf{s}^1$. Each level-1 score $\textbf{s}^1_i$ is the mean from $m$ under scores $\textbf{s}^2_j$ of level-2 tags, averaged from $k$ outputs $\textbf{o}_l$ (1 or 0) of a certain tag.

In addition to case-level and tag-level scores, we provide other analytical results, such as the distributions of response options to reflect model preferences and the invalid response rate out of A-E and corresponding keywords to assess format-following ability. Besides the averaged case scores, we also calculate the multiplied case scores to reflect the perfect cases UniScore, where all the image and questions are correct. All these metrics support holistic and in-depth analyses.

\vspace{-0.15cm}
\subsection{Human Evaluation}
\vspace{-0.1cm}
\label{sec_33}


\textbf{Setup}. To prove the proposed metric is highly consistent with human evaluation in visual generation, we conducted a human study and compared UniScore with other metrics \cite{hessel2021clipscore,li2024evaluating} for subjective visual generation. Conversely, understanding is an objective task with certain QAs, thus, the human study is not conducted as common practice. To focus on analyzing generation without being affected by the weak understanding capabilities of the unified models, we introduced Qwen2.5-VL-7B \cite{bai2025qwen2}  as the understanding model, which surpasses GPT-4v \cite{achiam2023gpt} on MMMU \cite{yue2024mmmu} (we also verified the 72B model in Appendix \ref{app_larger_model}). To ensure representativeness, we selected three models (Show-o \cite{xie2024show}, VILA-U \cite{wu2024vila}, Janus-Pro-7B \cite{chen2025janus}) and randomly sampled 100 different cases from UniBench for each, totaling 300 cases. We recruited three annotators with different educational backgrounds (PhD-annotator 0, UG-annotator 1, master-annotator 2) to conduct independent labeling, with annotations covering 3 annotators * 300 random cases * 4 images in total. The annotators were asked to label whether the keywords were expressed in the generated images for each questions. Given the complexity of labels and the subjectivity in generation, we provided four labels: (1) generation failure, (2) between success and failure, (3) successful generation, and (0) lacking knowledge to judge. Failures are scored as 0, successes as 1, while uncertain (score 2) and unknowing (score 4) labels are skipped. Both model results and human results associated with skipped labels are excluded from accuracy calculations to ensure the reliability of the evaluation (see human annotation examples in Appendix \ref{app_humaneval_case}).

\begin{figure}[htb]
  \centering
  \vspace{-0.15cm}
  \includegraphics[width=1\textwidth]{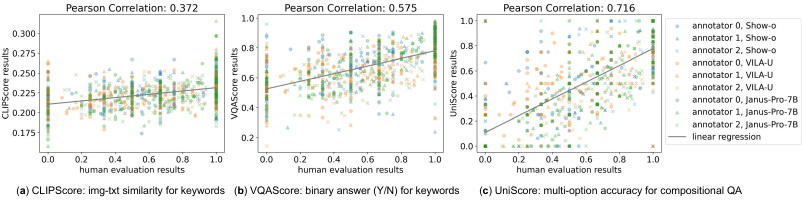}
  \vspace{-0.6cm}
  \caption{\label{fig_human_eval}\textbf{Correlation with Human Evaluation}. x-axis indicates the accuracy for a case from humans, scores in the y-axes are from CLIPScore \cite{hessel2021clipscore} (text-similarity), VQAScore \cite{li2024evaluating} (binary option confidence), and the proposed UniScore (multiple options accuracy). Pearson correlation \cite{cohen2009pearson} measures the normalized covariance, and a higher value indicates closer alignment.}
  \vspace{-0.15cm}
\end{figure}

\textbf{Analysis}. For each case, we calculate its accuracy as the human score on the x-axis of Fig. \ref{fig_human_eval}. The y-axis includes comparison metrics, including the commonly used instruction-following metric, CLIPScore \cite{hessel2021clipscore}, and the recent VQAScore. These two metrics represent auto-evaluation metrics based on text similarity and keyword-based binary options (existence or not), respectively. In contrast, we employ an accuracy-based metric with multiple options, which has smaller random errors with non-template questions (providing more information to avoid ambiguity). We use Pearson correlation \cite{cohen2009pearson} to assess the alignment between human and auto evaluations. Since Pearson correlation is a normalized covariance, it is independent of score distribution without any threshold. The results indicate that the proposed UniScore has an average correlation of 0.716 with human annotators, much higher than CLIPScore's 0.372 and VQAScore's 0.575, demonstrating a strong alignment with human judgment. We also provide more analyses in Appendix \ref{app_humaneval_ana} in aspects of annotators and models.


\vspace{-0.3cm}
\section{Experiments}
\vspace{-0.15cm}
\subsection{Implementation}
\vspace{-0.1cm}

\noindent\textbf{Evaluated Models.} For the unified setting, we have implemented ten open-source models via pytorch from their official codes and weights, including VARGPT \cite{zhuang2025vargpt}, TokenFlow \cite{qu2024tokenflow}, Show-o-Turbo \cite{xu2025show}, Show-o \cite{xie2024show}, Janus-Pro-1B \cite{chen2025janus}, Janus-1.3B \cite{wu2024janus}, VILA-U \cite{wu2024vila}, UniToken-stageII \cite{jiao2025unitoken}, JanusFlow-1.3B \cite{ma2024janusflow}, Janus-Pro-7B \cite{chen2025janus}. Some methods are not implemented due to unavailable models \cite{wang2024illume} or APIs \cite{gpt-4o}, non-pytorch environment \cite{liu2024world}, or dependence on third-party generation models \cite{dong2023dreamllm, ge2023planting, ge2023making, sunemu, ge2024seed, ye2024x, team2024chameleon}. We also released implementations of ten visual generation models in our codebase, including SDv1.5 \cite{rombach2022high}, SDv2.1 \cite{rombach2022high}, PixArt-$\alpha$ \cite{chen2023pixart}, SDXL \cite{podell2023sdxl}, FLUX.1-dev \cite{flux2024}, SDv3.5-Medium \cite{sd352025}, SDv3-Medium \cite{esser2024scaling}, FLUX.1-schnell \cite{flux2024}, DALL-E3 \cite{betker2023improving}, DALL-E2 \cite{ramesh2022hierarchical}.

\noindent\textbf{UniEval.} The implementation of UniBench has been described in Sec. \ref{sec_31}, including used LLMs (Deepseek-R1-70B \cite{guo2025deepseek}, QWQ-32B \cite{qwq32b}), Gaussain sampling hyperparameters (std \& mean = 6), with benchmark information (1,234 prompts and 4,231 QAs). The ``LLMs'' in step 1 involve POE, Gemini-1.5-Pro \cite{team2024gemini}, Deepseek-R1-70B, Qwen2.5-72B \cite{yang2024qwen2} on webs. More detailed prompts are shown in Appendix \ref{app_unibench_gen} with example cases in Appendix \ref{app_cases}. UniEval generates 4 images for each prompt, which are combined with each question to calculate the UniScore as described in Sec. \ref{sec_32}. When parsing responses, we prioritize matching letters A-E from a predefined format. If no match, use the last keyword found in the options. If neither exists, mark as invalid. Questions related to failed image generation also receive 0 scores. For details of human evaluation, like samples, models \cite{xu2025show,wu2024vila,chen2025janus,bai2025qwen2}, annotators, and compared metrics \cite{hessel2021clipscore,mathew2021docvqa}, are given in Sec. \ref{sec_33} with annotated cases in Appendix \ref{app_humaneval_case}.

\vspace{-0.15cm}
\subsection{Unified Multimodal Understanding and Generation}
\vspace{-0.15cm}
\noindent\textbf{Benchmarking Unified Models:} 
We implement most of the open-source unified models \cite{zhuang2025vargpt,qu2024tokenflow,xie2024show,xu2025show,chen2025janus,wu2024janus,wu2024vila,jiao2025unitoken,ma2024janusflow} and report their results on UniEval in Table \ref{tab_uni} as a leaderboard. We sort them by final UniScore and report the specific level-1 tag micro UniScores (see Appendix \ref{app_l2_res} for more detailed level-2 tag UniScores). Overall, the results range from 0.204 to 0.572, reflecting sufficient differences and difficulty. Compared to the random accuracy of about 0.5 from the binary choice benchmark \cite{ghosh2023geneval,cho2024davidsonian,huang2025t2i}, our benchmark significantly reduces the random error (expected 0.2). Tags such as adjectives, nouns, and styles perform well, while numerals, text, documents, and UI meet greater challenges. For specific models, VARGPT \cite{zhuang2025vargpt} performs worst, it only outputs texts instead of the required images for many prompts, resulting in 52\% invalid responses. The best performer is Janus-Pro-7B \cite{chen2025janus} at 0.572, whereas Janus-Pro-1B \cite{chen2025janus} performed worse than the earlier JanusFlow-1.3B \cite{ma2024janusflow}. This is due to Janus-Pro-1B's poor format-following, often failing to output required formats, leading to 16.7\% invalid responses. We discuss other anomalous results in detail in Sect. \ref{sec_insights} with corresponding insights.

\begin{table}[htbp]
  \vspace{-0.2cm}
  \caption{\label{tab_uni}\textbf{UniEval Results}. Fine-grained and overall UniScores on unified understanding and generation models. See tag names in Fig. \ref{fig_bench}a, level-2 scores in Appendix \ref{app_l2_res}, and analysis in Table \ref{tab_insights}.}
  \vspace{-0.1cm}
  \setlength{\tabcolsep}{3.3pt}
  \label{sample-table}
  \scriptsize
  \centering
  \begin{tabular}{ccccccccccccccc}
    \hline
    \textbf{Model} & \textbf{Num} & \textbf{Adj} & \textbf{Noun} & \textbf{Verb}  &  \textbf{Adv} &  \textbf{Text} &  \textbf{Doc}  &  \textbf{Sty} &  \textbf{Moda} &  \textbf{Qual} & \textbf{Effe} &  \textbf{Comp} &  \textbf{UI} & \textbf{UniScore}$\uparrow$\\
    \hline
    VARGPT \cite{zhuang2025vargpt} & 0.097 &0.326 &0.284 &0.288 &0.210 &0.049 &0.104 &0.227 &0.227 &0.335 &0.241 &0.155 &0.109 & 0.204 \\
    TokenFlow \cite{qu2024tokenflow} & 0.093 &0.522 &0.388 &0.330 &0.275 &0.157 &0.223 &0.600 &0.352 &0.163 &0.535 &0.517 &0.163 & 0.332 \\
    Show-o-Turbo \cite{xu2025show} & 0.250 &0.302 &0.353 &0.274 &0.256 &0.381 &0.331 &0.386 &0.331 &\textbf{0.546} &0.394 &0.360 &0.398 & 0.351 \\
    Show-o \cite{xie2024show} & 0.250 &0.362 &0.422 &0.316 &0.285 &0.381 &0.358 &0.390 &0.346 &0.472 &0.432 &0.360 &0.398 & 0.367 \\
    Janus-Pro-1B \cite{chen2025janus} & 0.186 &0.504 &0.443 &0.413 &0.370 &0.174 &0.233 &0.536 &0.503 &0.396 &0.397 &0.350 &0.301 & 0.370 \\
    Janus-1.3B \cite{wu2024janus} & 0.202 &0.484 &0.497 &0.384 &0.284 &0.246 &0.319 &0.641 &0.381 &0.408 &0.476 &0.423 &\underline{0.449} & 0.400 \\
    VILA-U \cite{wu2024vila} & 0.231 &0.604 &0.558 &0.549 &0.397 &0.254 &0.376 &0.704 &0.567 &0.362 &\underline{0.592} &0.453 &0.285 & 0.456 \\
    UniToken-II \cite{jiao2025unitoken} & \underline{0.349} &\underline{0.637} &\underline{0.624} &\underline{0.565} &0.386 &0.277 &\underline{0.430} &0.669 &0.593 &0.329 &0.568 &\textbf{0.589} &0.380 & 0.492 \\
    JanusFlow-1.3B \cite{ma2024janusflow} & 0.324 &0.608 &0.588 &0.528 &\underline{0.423} &\underline{0.400} &0.354 &\underline{0.706} &\underline{0.645} &0.521 &0.585 &0.496 &0.426 & \underline{0.508} \\
    Janus-Pro-7B \cite{chen2025janus} & \textbf{0.356} &\textbf{0.716} &\textbf{0.666} &\textbf{0.621} &\textbf{0.509} &\textbf{0.456} &\textbf{0.477} &\textbf{0.777} &\textbf{0.672} &\underline{0.542} &\textbf{0.655} &\underline{0.527} &\textbf{0.459} & \textbf{0.572} \\
    \hline
  \end{tabular}
  \vspace{-0.1cm}
\end{table}

\begin{wrapfigure}{r}{0.58\textwidth}
  \centering
  \vspace{-0.3cm}
  \includegraphics[width=0.58\textwidth]{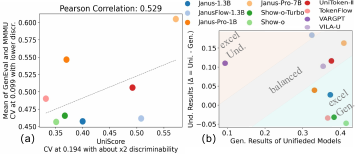}
  \vspace{-0.4cm}
  \caption{\label{fig_align_combine}\textbf{Task-specific Evaluation}. (\textbf{a}): UniEval aligns with the average of MMMU \cite{yue2024mmmu} and GenEval \cite{ghosh2023geneval}, exhibiting twice the discriminability measured by the coefficient of variation (CV). (\textbf{b}): UniEval supports task-specific evaluations. ``Gen. Results'' are evaluated with QWen2.5-VL-7B \cite{bai2025qwen2}. ``Und. Results'' are from the difference between Uni. and Gen. results (see Appendix \ref{app_specific_eval}), indicating preference on generation (blue region) or Und. in yellow.}
  \vspace{-0.2cm}
\end{wrapfigure}

\noindent\textbf{UniEval for Task-specific Evaluations:} 
First, we validated that UniEval has a relatively strong consistency (Pearson correlation of 0.529) with the combination of task-specific evaluations (representative MMMU \cite{yue2024mmmu} + GenEval \cite{ghosh2023geneval}) in Fig. \ref{fig_align_combine}a, demonstrating its rationality. We further calculated the coefficient of variation among models, showing that UniEval is twice that of task-specific evaluations, indicating a stronger discriminability (0.194 vs. 0.099). Analyzing outliers can reveal new insights from UniEval, such as the fact that Janus-Pro-1B \cite{wu2024janus} performs well independently, but its invalid response rate leads to poor unified results (see Table \ref{tab_insights}).
Second, UniEval also supports task-specific evaluations. As shown in Fig. \ref{fig_align_combine}b, we introduce QWen2.5-VL-7B \cite{bai2025qwen2} to obtain generation-only results. Then, use the difference between unified and generation-only results to analyze understanding capabilities. A higher score on the y-axis indicates that the unified model's understanding surpasses the extra understanding model. Among the models, JanusFlow-1.3B \cite{ma2024janusflow} demonstrates the best understanding of generated images, while many models struggled in understanding generated images, such as TokenFlow \cite{qu2024tokenflow}, Show-o \cite{xie2024show}, and Show-o-Turbo \cite{xu2025show}. These results indicate that UniEval not only excels in unified evaluation but also supports task-specific evaluation, enabling detailed model analysis for further improvements.

\begin{figure}[htb]
  \centering
  \includegraphics[width=1\textwidth]{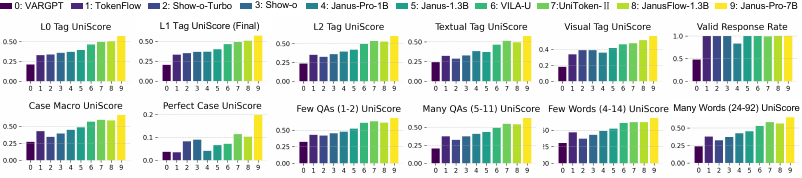}
  \vspace{-0.5cm}
  \caption{\label{fig_more_metrics}\textbf{Comparision with Detailed Metrics}. We illustrate results in more detailed metrics, including 5 tag-level scores on top with valid response rate, and 6 case-level scores on the bottom. Perfect indicates the all correct case ratio. Few and many are counted in varied QA and word sizes.}
  \vspace{-0.15cm}
\end{figure}

\noindent\textbf{Comparison in More Aspects:} 
We present various metrics in Fig. \ref{fig_more_metrics} for detailed analysis. The case macro UniScore and L1 tag UniScore align closely, but TokenFlow \cite{qu2024tokenflow} works better at the case level over the tag level, owing to weak abilities in some fine-grained tags like Num. and UI. The perfect case UniScore (all questions correct for 4 images in a case) is challenging, peaking at about 0.2, while the second-tier models are around 0.1. The valid response rate indicates generating expected texts or images as required, with obvious errors in VARGPT \cite{zhuang2025vargpt} and Janus-Pro-1B \cite{chen2025janus}. Model rankings for various word and QA sizes correspond with case scores, showing reduced results for larger sizes.

\begin{figure}[htb]
  \vspace{-0.1cm}
  \centering
  \includegraphics[width=1\textwidth]{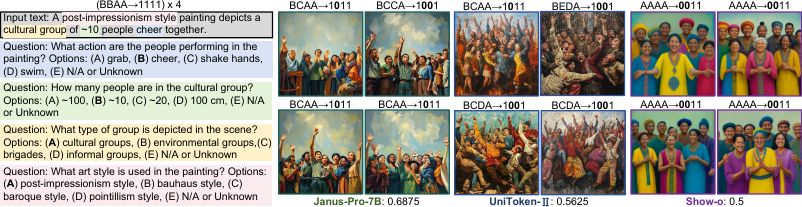}
  \vspace{-0.5cm}
  \caption{\label{fig_vis_case}\textbf{Visual Comparision}. An example with the responses of three unified models \cite{chen2025janus,jiao2025unitoken,xie2024show}. "BCAA→1011" indicates answers are BCAA of 4 questions, with the second one being incorrect.}
  \vspace{-0.15cm}
\end{figure}

\noindent\textbf{Case Studies:} 
In addition to the examples in Fig. \ref{fig_pipe}, we conducted a visual comparison of models in Fig. \ref{fig_vis_case}. For a case with four questions, we displayed the images generated by Janus-Pro-7B \cite{chen2025janus}, UniToken-II \cite{jiao2025unitoken}, and Show-o \cite{xie2024show}, along with their corresponding answers and correctness results. Janus-Pro-7B \cite{chen2025janus} achieved the highest case-level UniScore of 0.6875, while Show-o \cite{xie2024show} performed poorly due to biased responses (all A). Although UniToken-II produced images more similar to real images, it made obvious errors in visual generation regarding quantity and understanding of group types. This indicates that evaluation of instruction-following differs from image quality assessment, highlighting the unique value of UniEval. We also provide more case studies in Appendix \ref{app_cases}.

\vspace{-0.15cm}
\subsection{Text-to-Image Generation}
\vspace{-0.15cm}

\noindent\textbf{Benchmarking T2I Generation Models:} 

The proposed UniBench not only evaluates unified models but also supports visual generation models. Since generation-only models lack understanding capabilities, we introduced Qwen2.5-VL-7B \cite{bai2025qwen2} for automatic evaluation. This model outperforms GPT-4v \cite{achiam2023gpt} on MMMU \cite{yue2024mmmu} and aligns better with humans than the 72B model in Appendix \ref{app_larger_model}. We conducted extensive evaluations on 10 popular models in Table \ref{tab_gen}, including Stable Diffusion series \cite{rombach2022high,podell2023sdxl,sd352025,esser2024scaling}, PixArt-$\alpha$ \cite{chen2023pixart}, FLUX series \cite{flux2024}, and DALL-E series \cite{ramesh2022hierarchical, betker2023improving}. Results show that the earlier SDv1.5 \cite{rombach2022high} had the lowest UniScore of 0.33, while recent models like DALL-E \cite{betker2023improving}, FLUX \cite{flux2024}, and SDv3 \cite{esser2024scaling} performed well around 0.5. Instruction-following and image quality are not always aligned; for example, DALL-E3 is slightly lower than DALL-E2, because the prompt augmentation of DALL-E3 makes the prompt more complex with strict safety control. Similarly, FLUX.1-dev and SDv3.5-Medium sacrificed some instruction-following ability on this challenging benchmark. These insights are analyzed in Table \ref{tab_insights}. In Appendix \ref{app_specific_eval}, we compared UniScore for unified models using the same understanding model, showing that higher resolution benefits SoTA visual generation models in complex scenes, highlighting the need for higher resolution in unified models. Higher UniScores in Table \ref{tab_uni} indicate that unified models have a stronger understanding of generated images.

\begin{table}[htbp]
  \vspace{-0.2cm}
  \caption{\label{tab_gen}\textbf{UniBench for Visaul Generation}. UniBench is versatile and capable of evaluating text-to-image models. Note that measured instruction-following is different from image quality assessment, providing different insights. Bold and underline indicate the best and second, respectively.}
  \vspace{-0.1cm}
  \setlength{\tabcolsep}{3.1pt}
  \label{table_gen_only}
  \scriptsize
  \centering
  \begin{tabular}{ccccccccccccccc}
    \hline
    \textbf{Model} & \textbf{Num} & \textbf{Adj} & \textbf{Noun} & \textbf{Verb}  &  \textbf{Adv} &  \textbf{Text} &  \textbf{Doc}  &  \textbf{Sty} &  \textbf{Moda} &  \textbf{Qual} & \textbf{Effe} &  \textbf{Comp} &  \textbf{UI} & \textbf{UniScore}$\uparrow$\\
    \hline
    SDv1.5 \cite{rombach2022high} & 0.133 &0.453 &0.376 &0.274 &0.223 &0.065 &0.236 &0.673 &0.430 &0.238 &0.504 &0.505 &0.176 & 0.330 \\
    SDv2.1 \cite{rombach2022high} & 0.139 &0.479 &0.427 &0.305 &0.253 &0.101 &0.261 &0.678 &0.422 &0.266 &0.533 &0.519 &0.247 & 0.356 \\
    PixArt-$\alpha$ \cite{chen2023pixart} & 0.166 &0.550 &0.442 &0.348 &0.268 &0.065 &0.271 &0.729 &0.456 &0.233 &0.624 &0.616 &0.154 & 0.379 \\
    SDXL \cite{podell2023sdxl} & 0.149 &0.562 &0.461 &0.365 &0.311 &0.106 &0.294 &\underline{0.752} &0.512 &0.354 &0.626 &0.547 &0.210 & 0.404 \\
    FLUX.1-dev \cite{flux2024} & 0.270 &0.591 &0.470 &0.410 &0.321 &0.260 &0.459 &0.625 &0.453 &0.135 &0.589 &0.621 &0.307 & 0.424 \\
    SDv3.5-Medium \cite{sd352025} & 0.261 &0.609 &0.529 &0.421 &0.295 &0.318 &0.534 &0.718 &0.520 &0.346 &0.631 &0.547 &0.522 & 0.481 \\
    SDv3-Medium \cite{esser2024scaling} & 0.289 &0.581 &0.539 &0.461 &0.331 &0.378 &0.596 &0.670 &0.529 &0.314 &0.622 &0.568 &\underline{0.555} & 0.495 \\
    FLUX.1-schnell \cite{flux2024} & \underline{0.345} &\underline{0.642} &\underline{0.562} &0.451 &\underline{0.364} &0.305 &\underline{0.624} &0.717 &0.529 &0.190 &0.644 &0.606 &\textbf{0.644} & 0.509 \\
    DALL-E3 \cite{betker2023improving} & 0.312 &\textbf{0.650} &0.545 &\textbf{0.489} &\textbf{0.376} &\underline{0.375} &\textbf{0.627} &0.734 &\textbf{0.616} &\textbf{0.444} &\textbf{0.680} &\underline{0.632} &0.499 & \underline{0.537} \\
    DALL-E2 \cite{ramesh2022hierarchical} & \textbf{0.369} &0.624 &\textbf{0.605} &\underline{0.474} &0.360 &\underline{0.406} &0.610 &\textbf{0.762} &\underline{0.587} &\underline{0.362} &\underline{0.668} &\textbf{0.690} &0.527 & \textbf{0.542} \\
    \hline
  \end{tabular}
\end{table}

\begin{wrapfigure}{r}{0.5\textwidth}
  \centering
  \vspace{-0.3cm}
  \includegraphics[width=0.5\textwidth,height=0.22\textwidth]{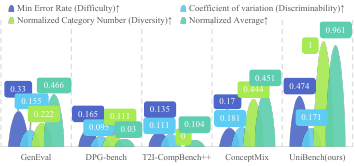}
  \vspace{-0.4cm}
  \caption{\label{fig_vs_bench}\textbf{Advantages Over T2I Benchmarks}. The proposed UniBench shows better difficulty and diversity beyond existing T2I benchmarks \cite{ghosh2023geneval,hu2024ella,huang2025t2i,wu2024conceptmix}. See specific data in Appendix \ref{app_comp_t2i}.}
  \vspace{-0.15cm}
\end{wrapfigure}

\noindent\textbf{Comparison with Other Benchmarks:} 
Our UniBench is not only the first unified benchmark, but it also outperforms existing benchmarks in evaluating text-to-image (T2I) models. In Fig. \ref{fig_vs_bench}, we compare T2I benchmarks from three perspectives: difficulty (error rate of the best model), discriminability (coefficient of variation), and diversity (min-max normalized number of attributes). For ConceptMix \cite{wu2024conceptmix}, we report its K=1 scores, as K>1 (multiplied accuracy) differs from others. Since T2I-CompBench++ uses multiple models to evaluate attributes and lacks an overall value, we report the complexity results based on GPT-4v \cite{achiam2023gpt} (see specific models and data in Appendix \ref{app_comp_t2i}). Experimental results show that UniBench significantly surpasses existing benchmarks in terms of difficulty and diversity, ranking second in discriminability (due to higher difficulty), with an overall score of 0.961, significantly exceeding the second, GenEval of 0.466.

\vspace{-0.15cm}
\subsection{Insights and Analysis}
\label{sec_insights}
\vspace{-0.15cm}
Studied from Table \ref{tab_uni}, Table \ref{tab_gen}, Fig. \ref{fig_align_combine}, Fig. \ref{fig_more_metrics}, and Appendix \ref{app_comp_t2i}, we conclude several interesting insights as shown in Table \ref{tab_insights} with cues and analyses. Key insights include: 1. Models like VARGPT \cite{zhuang2025vargpt} may fail to generate images (52\% invalid responses); 2. Models like Janus-Pro-1B \cite{chen2025janus} sometimes fail to follow the formats, with 16.7\% of responses outside A-E; 3. Show-o \cite{xie2024show} performs well with additional understanding models (Appendix \ref{app_comp_t2i}), but its own understanding model outputs very biased responses, with 89.2\% yielding A, leading to low unified results; 4. Models often struggle in some attributes, such as Num. (quantity and time); 5. UniBench introduced many new attributes, such as visual series labels; 6. Janus-Pro-7B \cite{chen2025janus} demonstrates good self-consistency and understanding of generated images, thus achieving a high UniScore; 7. Recent visual generation models yield higher UniScores beyond the unified model using the same understanding model because of higher resolution (1024 vs. 224-512), emphasizing the importance of resolution in complex prompts. 8. The extra models \cite{bai2025qwen2} tend to be stricter, generally scoring lower than unified models. 9. There is a trade-off between image quality and instruction-following; some earlier models like DALL-E2 \cite{ramesh2022hierarchical} may outperform newer models like DALL-E3 \cite{betker2023improving}, owing to the prompt augmentation in DALL-E3 (introducing extra content). Overall, UniEval provides valuable insights reflecting its unique values.

\vspace{-0.15cm}
\begin{table}[htbp]
  \caption{\label{tab_insights}\textbf{Insights}. UniEval provides valuable insights when evaluating unified and visual generation models, with the cues and corresponding analysis.}
    \vspace{-0.15cm}
  \setlength{\tabcolsep}{3pt}
  \label{sample-table}
  \scriptsize
  \centering
  \begin{tabular}{cccc}
    \hline
    \textbf{Insights} & \textbf{Model} & \textbf{Analysis} & \textbf{Cues}\\
    \hline
    Failure generation & VARGPT \cite{zhuang2025vargpt} & Often output texts when generating images  & 52\% invalid response\\
    Weak formatting & Janus-Pro-1B \cite{chen2025janus} & Often output response deviated from formats & 16.7\% responses out of ABCDE\\
    Biased response & Show-o \cite{xie2024show} & Ranks 1 in Appendix \ref{app_specific_eval} but only 7th in Table \ref{tab_uni} & 89.2\% responses in A\\
    Weak abilities & Uni. \& Gen. & Some tags are not generated well & e.g., quantity (Num)\\
    Visual aspects & Uni. \& Gen. & UniBench provides extensive visual tags & e.g.,  challenging web, language, and table\\
    Self-consistency & Janus-Pro-7B \cite{chen2025janus} & Good understanding for self-generated images & Only 7\% ``N/A or Unknown''\\
    Crucial resolution & Uni. vs. Gen. & Higer resolutions bring higher UniScore & Table \ref{tab_gen} vs. Appendix \ref{app_specific_eval} \\
    Strict criteria & Qwen-2.5VL-7B \cite{bai2025qwen2} & Extra model is strict and may reduce UniScore & More ``N/A or Unknown''\\
    Trade-off & Gen. models & Better quality may not enhance instruction-following & e.g., DALL-E2 vs. DALL-E3\\
    \hline
  \end{tabular}
  \vspace{-0.2cm}
\end{table}

\vspace{-0.2cm}
\section{Conclusion}
\vspace{-0.2cm}
In conclusion, we proposed UniEval, the first evaluation framework for unified multimodal understanding and generation models. By addressing the limitations of existing task-specific benchmarks, UniEval eliminates the reliance on extra models and images, enabling an elegant, simplified, and overall evaluation. The involved benchmark, UniBench, provides the most holistic fine-grained attributes beyond existing generation benchmarks, with the UniScore metric aligned well with human evaluations. Through extensive evaluations on SoTA unified and visual generation models, UniEval offers many valuable insights reflecting its unique values. As the field continues to evolve, UniEval stands out as a pioneering tool that can foster further advancements in multimodal understanding and generation. As the first unified evaluation framework, UniEval sets a new standard for unified multimodal evaluation and well supports visual generation fields, fostering continued progress in generative AI. Besides the significances of UniEval, we also discussed the limitations in Appendix \ref{app_limitaion}.

\clearpage
{
    \small
    \bibliographystyle{plain}
    \bibliography{main}

\begin{thebibliography}{10}

\bibitem{achiam2023gpt}
Josh Achiam, Steven Adler, Sandhini Agarwal, Lama Ahmad, Ilge Akkaya, Florencia~Leoni Aleman, Diogo Almeida, Janko Altenschmidt, Sam Altman, Shyamal Anadkat, et~al.
\newblock Gpt-4 technical report.
\newblock {\em arXiv preprint arXiv:2303.08774}, 2023.

\bibitem{antol2015vqa}
Stanislaw Antol, Aishwarya Agrawal, Jiasen Lu, Margaret Mitchell, Dhruv Batra, C~Lawrence Zitnick, and Devi Parikh.
\newblock Vqa: Visual question answering.
\newblock In {\em Proceedings of the IEEE international conference on computer vision}, pages 2425--2433, 2015.

\bibitem{bai2025qwen2}
Shuai Bai, Keqin Chen, Xuejing Liu, Jialin Wang, Wenbin Ge, Sibo Song, Kai Dang, Peng Wang, Shijie Wang, Jun Tang, et~al.
\newblock Qwen2. 5-vl technical report.
\newblock {\em arXiv preprint arXiv:2502.13923}, 2025.

\bibitem{bakr2023hrs}
Eslam~Mohamed Bakr, Pengzhan Sun, Xiaoqian Shen, Faizan~Farooq Khan, Li~Erran Li, and Mohamed Elhoseiny.
\newblock Hrs-bench: Holistic, reliable and scalable benchmark for text-to-image models.
\newblock In {\em Proceedings of the IEEE/CVF International Conference on Computer Vision}, pages 20041--20053, 2023.

\bibitem{barratt2018note}
Shane Barratt and Rishi Sharma.
\newblock A note on the inception score.
\newblock {\em arXiv preprint arXiv:1801.01973}, 2018.

\bibitem{betker2023improving}
James Betker, Gabriel Goh, Li~Jing, Tim Brooks, Jianfeng Wang, Linjie Li, Long Ouyang, Juntang Zhuang, Joyce Lee, Yufei Guo, et~al.
\newblock Improving image generation with better captions.
\newblock {\em Computer Science. https://cdn. openai. com/papers/dall-e-3. pdf}, 2(3):8, 2023.

\bibitem{chen2023pixart}
Junsong Chen, Jincheng Yu, Chongjian Ge, Lewei Yao, Enze Xie, Yue Wu, Zhongdao Wang, James Kwok, Ping Luo, Huchuan Lu, et~al.
\newblock Pixart-alpha: Fast training of diffusion transformer for photorealistic text-to-image synthesis.
\newblock {\em arXiv preprint arXiv:2310.00426}, 2023.

\bibitem{chen2025janus}
Xiaokang Chen, Zhiyu Wu, Xingchao Liu, Zizheng Pan, Wen Liu, Zhenda Xie, Xingkai Yu, and Chong Ruan.
\newblock Janus-pro: Unified multimodal understanding and generation with data and model scaling.
\newblock {\em arXiv preprint arXiv:2501.17811}, 2025.

\bibitem{chen2015microsoft}
Xinlei Chen, Hao Fang, Tsung-Yi Lin, Ramakrishna Vedantam, Saurabh Gupta, Piotr Doll{\'a}r, and C~Lawrence Zitnick.
\newblock Microsoft coco captions: Data collection and evaluation server.
\newblock {\em arXiv preprint arXiv:1504.00325}, 2015.

\bibitem{chen2024internvl}
Zhe Chen, Jiannan Wu, Wenhai Wang, Weijie Su, Guo Chen, Sen Xing, Muyan Zhong, Qinglong Zhang, Xizhou Zhu, Lewei Lu, et~al.
\newblock Internvl: Scaling up vision foundation models and aligning for generic visual-linguistic tasks.
\newblock In {\em Proceedings of the IEEE/CVF conference on computer vision and pattern recognition}, pages 24185--24198, 2024.

\bibitem{cho2024davidsonian}
Jaemin Cho, Yushi Hu, Jason~M Baldridge, Roopal Garg, Peter Anderson, Ranjay Krishna, Mohit Bansal, Jordi Pont-Tuset, and Su~Wang.
\newblock Davidsonian scene graph: Improving reliability in fine-grained evaluation for text-to-image generation.
\newblock In {\em ICLR}, 2024.

\bibitem{cohen2009pearson}
Israel Cohen, Yiteng Huang, Jingdong Chen, Jacob Benesty, Jacob Benesty, Jingdong Chen, Yiteng Huang, and Israel Cohen.
\newblock Pearson correlation coefficient.
\newblock {\em Noise reduction in speech processing}, pages 1--4, 2009.

\bibitem{deng2009imagenet}
Jia Deng, Wei Dong, Richard Socher, Li-Jia Li, Kai Li, and Li~Fei-Fei.
\newblock Imagenet: A large-scale hierarchical image database.
\newblock In {\em 2009 IEEE conference on computer vision and pattern recognition}, pages 248--255. Ieee, 2009.

\bibitem{denkowski2011meteor}
Michael Denkowski and Alon Lavie.
\newblock Meteor 1.3: Automatic metric for reliable optimization and evaluation of machine translation systems.
\newblock In {\em Proceedings of the sixth workshop on statistical machine translation}, pages 85--91, 2011.

\bibitem{dong2023dreamllm}
Runpei Dong, Chunrui Han, Yuang Peng, Zekun Qi, Zheng Ge, Jinrong Yang, Liang Zhao, Jianjian Sun, Hongyu Zhou, Haoran Wei, et~al.
\newblock Dreamllm: Synergistic multimodal comprehension and creation.
\newblock {\em arXiv preprint arXiv:2309.11499}, 2023.

\bibitem{esser2024scaling}
Patrick Esser, Sumith Kulal, Andreas Blattmann, Rahim Entezari, Jonas M{\"u}ller, Harry Saini, Yam Levi, Dominik Lorenz, Axel Sauer, Frederic Boesel, et~al.
\newblock Scaling rectified flow transformers for high-resolution image synthesis.
\newblock In {\em Forty-first international conference on machine learning}, 2024.

\bibitem{fengtraining}
Weixi Feng, Xuehai He, Tsu-Jui Fu, Varun Jampani, Arjun~Reddy Akula, Pradyumna Narayana, Sugato Basu, Xin~Eric Wang, and William~Yang Wang.
\newblock Training-free structured diffusion guidance for compositional text-to-image synthesis.
\newblock In {\em The Eleventh International Conference on Learning Representations}.

\bibitem{ge2023planting}
Yuying Ge, Yixiao Ge, Ziyun Zeng, Xintao Wang, and Ying Shan.
\newblock Planting a seed of vision in large language model.
\newblock {\em arXiv preprint arXiv:2307.08041}, 2023.

\bibitem{ge2023making}
Yuying Ge, Sijie Zhao, Ziyun Zeng, Yixiao Ge, Chen Li, Xintao Wang, and Ying Shan.
\newblock Making llama see and draw with seed tokenizer.
\newblock {\em arXiv preprint arXiv:2310.01218}, 2023.

\bibitem{ge2024seed}
Yuying Ge, Sijie Zhao, Jinguo Zhu, Yixiao Ge, Kun Yi, Lin Song, Chen Li, Xiaohan Ding, and Ying Shan.
\newblock Seed-x: Multimodal models with unified multi-granularity comprehension and generation.
\newblock {\em arXiv preprint arXiv:2404.14396}, 2024.

\bibitem{ghosh2023geneval}
Dhruba Ghosh, Hannaneh Hajishirzi, and Ludwig Schmidt.
\newblock Geneval: An object-focused framework for evaluating text-to-image alignment.
\newblock {\em Advances in Neural Information Processing Systems}, 36:52132--52152, 2023.

\bibitem{guo2025deepseek}
Daya Guo, Dejian Yang, Haowei Zhang, Junxiao Song, Ruoyu Zhang, Runxin Xu, Qihao Zhu, Shirong Ma, Peiyi Wang, Xiao Bi, et~al.
\newblock Deepseek-r1: Incentivizing reasoning capability in llms via reinforcement learning.
\newblock {\em arXiv preprint arXiv:2501.12948}, 2025.

\bibitem{hessel2021clipscore}
Jack Hessel, Ari Holtzman, Maxwell Forbes, Ronan Le~Bras, and Yejin Choi.
\newblock Clipscore: A reference-free evaluation metric for image captioning.
\newblock In {\em EMNLP (1)}, 2021.

\bibitem{heusel2017gans}
Martin Heusel, Hubert Ramsauer, Thomas Unterthiner, Bernhard Nessler, and Sepp Hochreiter.
\newblock Gans trained by a two time-scale update rule converge to a local nash equilibrium.
\newblock {\em Advances in neural information processing systems}, 30, 2017.

\bibitem{higgins2017scan}
Irina Higgins, Nicolas Sonnerat, Loic Matthey, Arka Pal, Christopher~P Burgess, Matko Bosnjak, Murray Shanahan, Matthew Botvinick, Demis Hassabis, and Alexander Lerchner.
\newblock Scan: Learning hierarchical compositional visual concepts.
\newblock {\em arXiv preprint arXiv:1707.03389}, 2017.

\bibitem{hu2024ella}
Xiwei Hu, Rui Wang, Yixiao Fang, Bin Fu, Pei Cheng, and Gang Yu.
\newblock Ella: Equip diffusion models with llm for enhanced semantic alignment.
\newblock {\em arXiv preprint arXiv:2403.05135}, 2024.

\bibitem{huang2025t2i}
Kaiyi Huang, Chengqi Duan, Kaiyue Sun, Enze Xie, Zhenguo Li, and Xihui Liu.
\newblock T2i-compbench++: An enhanced and comprehensive benchmark for compositional text-to-image generation.
\newblock {\em IEEE Transactions on Pattern Analysis and Machine Intelligence}, 2025.

\bibitem{hudson2019gqa}
Drew~A Hudson and Christopher~D Manning.
\newblock Gqa: A new dataset for real-world visual reasoning and compositional question answering.
\newblock In {\em Proceedings of the IEEE/CVF conference on computer vision and pattern recognition}, pages 6700--6709, 2019.

\bibitem{hurst2024gpt}
Aaron Hurst, Adam Lerer, Adam~P Goucher, Adam Perelman, Aditya Ramesh, Aidan Clark, AJ~Ostrow, Akila Welihinda, Alan Hayes, Alec Radford, et~al.
\newblock Gpt-4o system card.
\newblock {\em arXiv preprint arXiv:2410.21276}, 2024.

\bibitem{jiao2025unitoken}
Yang Jiao, Haibo Qiu, Zequn Jie, Shaoxiang Chen, Jingjing Chen, Lin Ma, and Yu-Gang Jiang.
\newblock Unitoken: Harmonizing multimodal understanding and generation through unified visual encoding.
\newblock {\em arXiv preprint arXiv:2504.04423}, 2025.

\bibitem{flux2024}
Black~Forest Labs.
\newblock Flux.
\newblock \url{https://github.com/black-forest-labs/flux}, 2024.

\bibitem{lee2023holistic}
Tony Lee, Michihiro Yasunaga, Chenlin Meng, Yifan Mai, Joon~Sung Park, Agrim Gupta, Yunzhi Zhang, Deepak Narayanan, Hannah Teufel, Marco Bellagente, et~al.
\newblock Holistic evaluation of text-to-image models.
\newblock {\em Advances in Neural Information Processing Systems}, 36:69981--70011, 2023.

\bibitem{li2024evaluating}
Baiqi Li, Zhiqiu Lin, Deepak Pathak, Jiayao Li, Yixin Fei, Kewen Wu, Xide Xia, Pengchuan Zhang, Graham Neubig, and Deva Ramanan.
\newblock Evaluating and improving compositional text-to-visual generation.
\newblock In {\em Proceedings of the IEEE/CVF Conference on Computer Vision and Pattern Recognition}, pages 5290--5301, 2024.

\bibitem{li2023seed}
Bohao Li, Rui Wang, Guangzhi Wang, Yuying Ge, Yixiao Ge, and Ying Shan.
\newblock Seed-bench: Benchmarking multimodal llms with generative comprehension.
\newblock {\em arXiv preprint arXiv:2307.16125}, 2023.

\bibitem{li2024dual}
Zijie Li, Henry Li, Yichun Shi, Amir~Barati Farimani, Yuval Kluger, Linjie Yang, and Peng Wang.
\newblock Dual diffusion for unified image generation and understanding.
\newblock {\em arXiv preprint arXiv:2501.00289}, 2024.

\bibitem{lin2014microsoft}
Tsung-Yi Lin, Michael Maire, Serge Belongie, James Hays, Pietro Perona, Deva Ramanan, Piotr Doll{\'a}r, and C~Lawrence Zitnick.
\newblock Microsoft coco: Common objects in context.
\newblock In {\em Computer vision--ECCV 2014: 13th European conference, zurich, Switzerland, September 6-12, 2014, proceedings, part v 13}, pages 740--755. Springer, 2014.

\bibitem{liu2024world}
Hao Liu, Wilson Yan, Matei Zaharia, and Pieter Abbeel.
\newblock World model on million-length video and language with ringattention.
\newblock {\em arXiv e-prints}, pages arXiv--2402, 2024.

\bibitem{liu2023visual}
Haotian Liu, Chunyuan Li, Qingyang Wu, and Yong~Jae Lee.
\newblock Visual instruction tuning.
\newblock {\em Advances in neural information processing systems}, 36:34892--34916, 2023.

\bibitem{liu2024mmbench}
Yuan Liu, Haodong Duan, Yuanhan Zhang, Bo~Li, Songyang Zhang, Wangbo Zhao, Yike Yuan, Jiaqi Wang, Conghui He, Ziwei Liu, et~al.
\newblock Mmbench: Is your multi-modal model an all-around player?
\newblock In {\em European conference on computer vision}, pages 216--233. Springer, 2024.

\bibitem{lu2022learn}
Pan Lu, Swaroop Mishra, Tanglin Xia, Liang Qiu, Kai-Wei Chang, Song-Chun Zhu, Oyvind Tafjord, Peter Clark, and Ashwin Kalyan.
\newblock Learn to explain: Multimodal reasoning via thought chains for science question answering.
\newblock {\em Advances in Neural Information Processing Systems}, 35:2507--2521, 2022.

\bibitem{ma2024janusflow}
Yiyang Ma, Xingchao Liu, Xiaokang Chen, Wen Liu, Chengyue Wu, Zhiyu Wu, Zizheng Pan, Zhenda Xie, Haowei Zhang, Liang Zhao, et~al.
\newblock Janusflow: Harmonizing autoregression and rectified flow for unified multimodal understanding and generation.
\newblock {\em arXiv preprint arXiv:2411.07975}, 2024.

\bibitem{mathew2021docvqa}
Minesh Mathew, Dimosthenis Karatzas, and CV~Jawahar.
\newblock Docvqa: A dataset for vqa on document images.
\newblock In {\em Proceedings of the IEEE/CVF winter conference on applications of computer vision}, pages 2200--2209, 2021.

\bibitem{mishra2019ocr}
Anand Mishra, Shashank Shekhar, Ajeet~Kumar Singh, and Anirban Chakraborty.
\newblock Ocr-vqa: Visual question answering by reading text in images.
\newblock In {\em 2019 international conference on document analysis and recognition (ICDAR)}, pages 947--952. IEEE, 2019.

\bibitem{gpt-4o}
OpenAI.
\newblock Gpt-4o.
\newblock \url{https://openai.com/index/hello-gpt-4o/}, 2025.

\bibitem{papineni2002bleu}
Kishore Papineni et~al.
\newblock Bleu: a method for automatic evaluation of machine translation.
\newblock In {\em Proceedings of the 40th annual meeting of the Association for Computational Linguistics}, pages 311--318, 2002.

\bibitem{peng2024dreambench++}
Yuang Peng, Yuxin Cui, Haomiao Tang, Zekun Qi, Runpei Dong, Jing Bai, Chunrui Han, Zheng Ge, Xiangyu Zhang, and Shu-Tao Xia.
\newblock Dreambench++: A human-aligned benchmark for personalized image generation.
\newblock {\em arXiv preprint arXiv:2406.16855}, 2024.

\bibitem{podell2023sdxl}
Dustin Podell, Zion English, Kyle Lacey, Andreas Blattmann, Tim Dockhorn, Jonas M{\"u}ller, Joe Penna, and Robin Rombach.
\newblock Sdxl: Improving latent diffusion models for high-resolution image synthesis.
\newblock {\em arXiv preprint arXiv:2307.01952}, 2023.

\bibitem{qu2024tokenflow}
Liao Qu, Huichao Zhang, Yiheng Liu, Xu~Wang, Yi~Jiang, Yiming Gao, Hu~Ye, Daniel~K Du, Zehuan Yuan, and Xinglong Wu.
\newblock Tokenflow: Unified image tokenizer for multimodal understanding and generation.
\newblock {\em arXiv preprint arXiv:2412.03069}, 2024.

\bibitem{ramesh2022hierarchical}
Aditya Ramesh, Prafulla Dhariwal, Alex Nichol, Casey Chu, and Mark Chen.
\newblock Hierarchical text-conditional image generation with clip latents.
\newblock {\em arXiv preprint arXiv:2204.06125}, 1(2):3, 2022.

\bibitem{rombach2022high}
Robin Rombach, Andreas Blattmann, Dominik Lorenz, Patrick Esser, and Bj{\"o}rn Ommer.
\newblock High-resolution image synthesis with latent diffusion models.
\newblock In {\em Proceedings of the IEEE/CVF conference on computer vision and pattern recognition}, pages 10684--10695, 2022.

\bibitem{saharia2022photorealistic}
Chitwan Saharia, William Chan, Saurabh Saxena, Lala Li, Jay Whang, Emily~L Denton, Kamyar Ghasemipour, Raphael Gontijo~Lopes, Burcu Karagol~Ayan, Tim Salimans, et~al.
\newblock Photorealistic text-to-image diffusion models with deep language understanding.
\newblock {\em Advances in neural information processing systems}, 35:36479--36494, 2022.

\bibitem{salimans2016improved}
Tim Salimans, Ian Goodfellow, Wojciech Zaremba, Vicki Cheung, Alec Radford, and Xi~Chen.
\newblock Improved techniques for training gans.
\newblock {\em Advances in neural information processing systems}, 29, 2016.

\bibitem{schuhmann2022laion}
Christoph Schuhmann, Romain Beaumont, Richard Vencu, Cade Gordon, Ross Wightman, Mehdi Cherti, Theo Coombes, Aarush Katta, Clayton Mullis, Mitchell Wortsman, et~al.
\newblock Laion-5b: An open large-scale dataset for training next generation image-text models.
\newblock {\em Advances in neural information processing systems}, 35:25278--25294, 2022.

\bibitem{shah2019kvqa}
Sanket Shah, Anand Mishra, Naganand Yadati, and Partha~Pratim Talukdar.
\newblock Kvqa: Knowledge-aware visual question answering.
\newblock In {\em Proceedings of the AAAI conference on artificial intelligence}, volume~33, pages 8876--8884, 2019.

\bibitem{sd352025}
stability.ai.
\newblock Sdv3.5.
\newblock \url{https://stability.ai/news/introducing-stable-diffusion-3-5}, 2025.

\bibitem{sunemu}
Quan Sun, Qiying Yu, Yufeng Cui, Fan Zhang, Xiaosong Zhang, Yueze Wang, Hongcheng Gao, Jingjing Liu, Tiejun Huang, and Xinlong Wang.
\newblock Emu: Generative pretraining in multimodality.
\newblock In {\em The Twelfth International Conference on Learning Representations}.

\bibitem{team2024chameleon}
Chameleon Team.
\newblock Chameleon: Mixed-modal early-fusion foundation models.
\newblock {\em arXiv preprint arXiv:2405.09818}, 2024.

\bibitem{team2024gemini}
Gemini Team, Petko Georgiev, Ving~Ian Lei, Ryan Burnell, Libin Bai, Anmol Gulati, Garrett Tanzer, Damien Vincent, Zhufeng Pan, Shibo Wang, et~al.
\newblock Gemini 1.5: Unlocking multimodal understanding across millions of tokens of context.
\newblock {\em arXiv preprint arXiv:2403.05530}, 2024.

\bibitem{qwq32b}
Qwen Team.
\newblock Qwq-32b: Embracing the power of reinforcement learning, March 2025.

\bibitem{tian2025quality}
Yu~Tian, Yue Liu, Shiqi Wang, and Sam Kwong.
\newblock Quality assessment for text-to-image generation: A survey.
\newblock {\em IEEE MultiMedia}, 2025.

\bibitem{wang2024illume}
Chunwei Wang, Guansong Lu, Junwei Yang, Runhui Huang, Jianhua Han, Lu~Hou, Wei Zhang, and Hang Xu.
\newblock Illume: Illuminating your llms to see, draw, and self-enhance.
\newblock {\em arXiv preprint arXiv:2412.06673}, 2024.

\bibitem{wu2024janus}
Chengyue Wu, Xiaokang Chen, Zhiyu Wu, Yiyang Ma, Xingchao Liu, Zizheng Pan, Wen Liu, Zhenda Xie, Xingkai Yu, Chong Ruan, et~al.
\newblock Janus: Decoupling visual encoding for unified multimodal understanding and generation.
\newblock {\em arXiv preprint arXiv:2410.13848}, 2024.

\bibitem{wu2024conceptmix}
Xindi Wu, Dingli Yu, Yangsibo Huang, Olga Russakovsky, and Sanjeev Arora.
\newblock Conceptmix: A compositional image generation benchmark with controllable difficulty.
\newblock {\em arXiv preprint arXiv:2408.14339}, 2024.

\bibitem{wu2024vila}
Yecheng Wu, Zhuoyang Zhang, Junyu Chen, Haotian Tang, Dacheng Li, Yunhao Fang, Ligeng Zhu, Enze Xie, Hongxu Yin, Li~Yi, et~al.
\newblock Vila-u: a unified foundation model integrating visual understanding and generation.
\newblock {\em arXiv preprint arXiv:2409.04429}, 2024.

\bibitem{xie2024show}
Jinheng Xie, Weijia Mao, Zechen Bai, David~Junhao Zhang, Weihao Wang, Kevin~Qinghong Lin, Yuchao Gu, Zhijie Chen, Zhenheng Yang, and Mike~Zheng Shou.
\newblock Show-o: One single transformer to unify multimodal understanding and generation.
\newblock {\em arXiv preprint arXiv:2408.12528}, 2024.

\bibitem{xu2025show}
Chenkai Xu, Xu~Wang, Zhenyi Liao, Yishun Li, Tianqi Hou, and Zhijie Deng.
\newblock Show-o turbo: Towards accelerated unified multimodal understanding and generation.
\newblock {\em arXiv preprint arXiv:2502.05415}, 2025.

\bibitem{xu2023imagereward}
Jiazheng Xu, Xiao Liu, Yuchen Wu, Yuxuan Tong, Qinkai Li, Ming Ding, Jie Tang, and Yuxiao Dong.
\newblock Imagereward: Learning and evaluating human preferences for text-to-image generation.
\newblock {\em Advances in Neural Information Processing Systems}, 36:15903--15935, 2023.

\bibitem{yang2024qwen2}
An~Yang, Baosong Yang, Beichen Zhang, Binyuan Hui, Bo~Zheng, Bowen Yu, Chengyuan Li, Dayiheng Liu, Fei Huang, Haoran Wei, et~al.
\newblock Qwen2. 5 technical report.
\newblock {\em arXiv preprint arXiv:2412.15115}, 2024.

\bibitem{ye2024x}
Hanrong Ye, De-An Huang, Yao Lu, Zhiding Yu, Wei Ping, Andrew Tao, Jan Kautz, Song Han, Dan Xu, Pavlo Molchanov, et~al.
\newblock X-vila: Cross-modality alignment for large language model.
\newblock {\em arXiv preprint arXiv:2405.19335}, 2024.

\bibitem{ye2023mplug}
Qinghao Ye, Haiyang Xu, Guohai Xu, Jiabo Ye, Ming Yan, Yiyang Zhou, Junyang Wang, Anwen Hu, Pengcheng Shi, Yaya Shi, et~al.
\newblock mplug-owl: Modularization empowers large language models with multimodality.
\newblock {\em arXiv preprint arXiv:2304.14178}, 2023.

\bibitem{yue2024mmmu}
Xiang Yue, Yuansheng Ni, Kai Zhang, Tianyu Zheng, Ruoqi Liu, Ge~Zhang, Samuel Stevens, Dongfu Jiang, Weiming Ren, Yuxuan Sun, et~al.
\newblock Mmmu: A massive multi-discipline multimodal understanding and reasoning benchmark for expert agi.
\newblock In {\em Proceedings of the IEEE/CVF Conference on Computer Vision and Pattern Recognition}, pages 9556--9567, 2024.

\bibitem{zhou2024transfusion}
Chunting Zhou, Lili Yu, Arun Babu, Kushal Tirumala, Michihiro Yasunaga, Leonid Shamis, Jacob Kahn, Xuezhe Ma, Luke Zettlemoyer, and Omer Levy.
\newblock Transfusion: Predict the next token and diffuse images with one multi-modal model.
\newblock {\em arXiv preprint arXiv:2408.11039}, 2024.

\bibitem{zhuang2025vargpt}
Xianwei Zhuang, Yuxin Xie, Yufan Deng, Liming Liang, Jinghan Ru, Yuguo Yin, and Yuexian Zou.
\newblock Vargpt: Unified understanding and generation in a visual autoregressive multimodal large language model.
\newblock {\em arXiv preprint arXiv:2501.12327}, 2025.

\end{thebibliography}
}
\clearpage

\appendix

\begin{center}
\textbf{\LARGE Appendix}
\end{center}

\tableofcontents
\clearpage

\section{Results Comparision Among Level-2 Tags}
\label{app_l2_res}
\begin{figure}[H]
  \centering
  \includegraphics[width=1\textwidth]{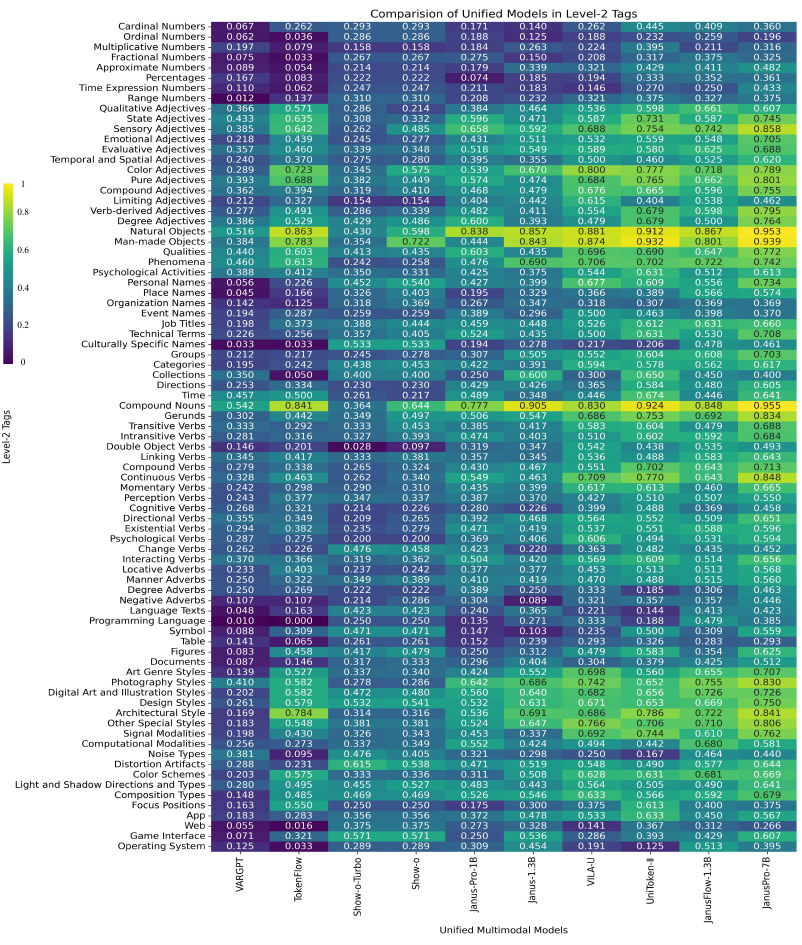}
  \vspace{-0.2cm}
  \caption{\label{fig_l2_score_uni}Results of \textbf{Unified Models} Evaluated on Level-2 Tags.}
\end{figure}

Besides level-1 results in Table \ref{tab_uni} and Table \ref{tab_gen}, we present more detailed results for level-2 tags in this section. Fig. \ref{fig_l2_score_uni} compares the unified multimodal models \cite{zhuang2025vargpt,qu2024tokenflow,xu2025show,xie2024show,chen2025janus,wu2024janus,wu2024vila,jiao2025unitoken,ma2024janusflow}, where results with yellow backgrounds indicate better-performing attributes. Overall, adjectives, nouns, and style-related level-2 tags perform well, such as natural objects, man-made objects, and compound nouns, achieving many results above 0.9. Numbers, texts, UI, and other related level-2 tags are more challenging, with many models performing below 0.3, and TokenFlow \cite{qu2024tokenflow} even scoring 0 in the programming language. From the model view, Janus-Pro-7B \cite{chen2025janus} performs the best, while VARGPT \cite{zhuang2025vargpt} performs the worst.

\begin{figure}[H]
  \centering
  \vspace{-0.5cm}
  \includegraphics[width=1\textwidth]{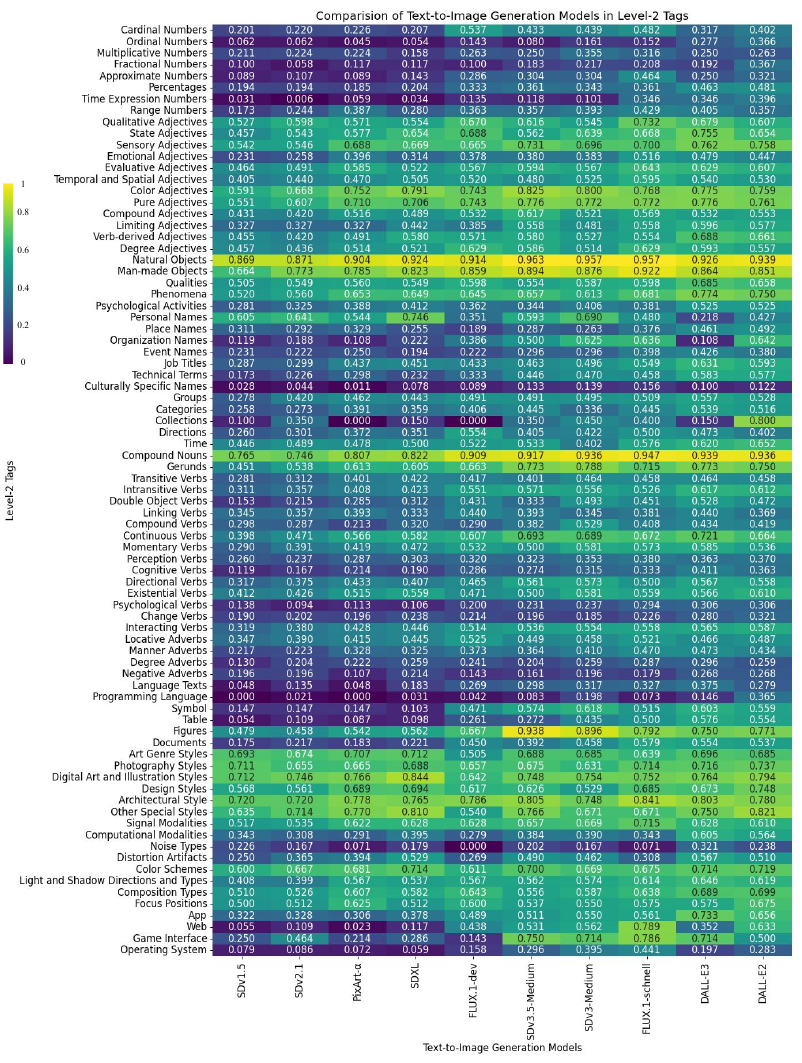}
  \vspace{-0.2cm}
  \caption{\label{fig_l2_score_gen}Results of \textbf{Visual Generation Models} Evaluated on Level-2 Tags.}
\end{figure}

Fig. \ref{fig_l2_score_gen} compares visual generation models \cite{rombach2022high,chen2023pixart,podell2023sdxl,flux2024,sd352025,esser2024scaling,betker2023improving,ramesh2022hierarchical}, and the overall performances are similar to those of the unified models, but the differences between high and low performances are more pronounced. For example, natural objects have a minimum score of 0.869, significantly higher than the unified models' lowest score of 0.43. In contrast, for culturally specific names that require reasoning, the highest score for generation-only models is only 0.156, whereas the unified models reach a maximum of 0.533. This indicates that visual generation models perform better for common attributes (benefiting from higher resolution), while unified models excel in instruction-following under complex conditions.

\section{Examples of Keywords}
\begin{figure}[H]
  \centering
  \vspace{-0.2cm}
  \includegraphics[width=1\textwidth]{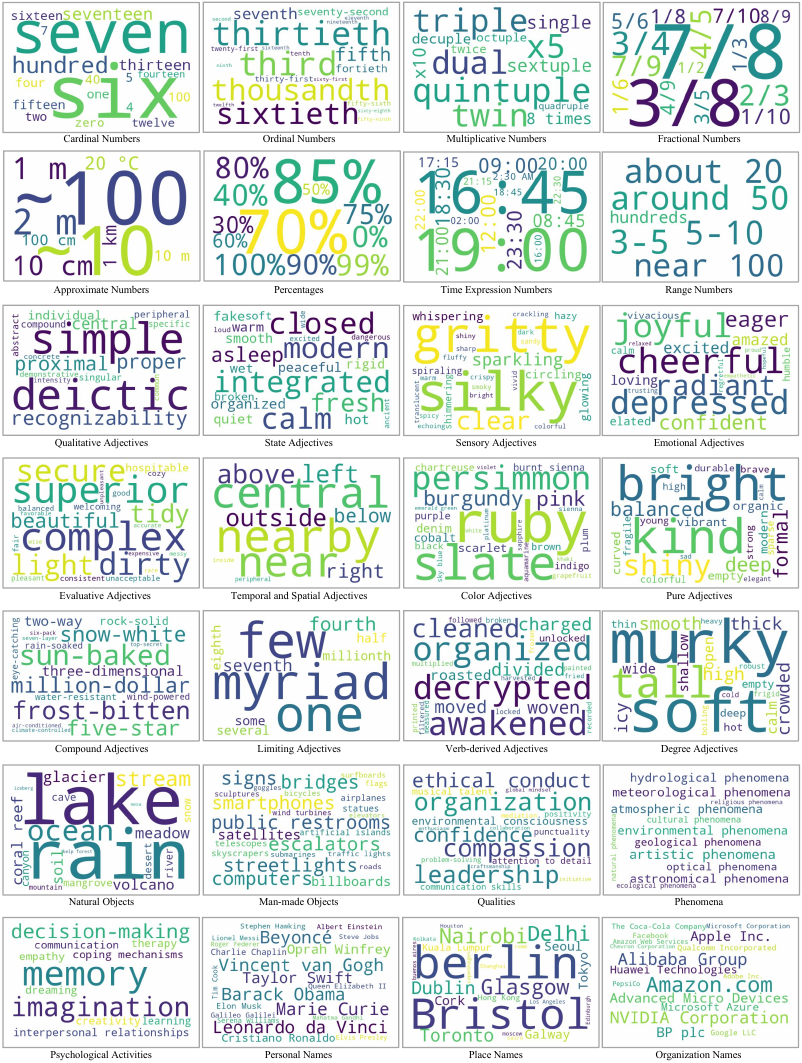}
  \vspace{-0.4cm}
  \caption{\label{fig_keywords1}\textbf{Example of keywords}. For each level-2 tag, we displayed up to 15 keywords through the word cloud, with random sampling and text size.}
\end{figure}

\begin{figure}[H]
  \centering
  \includegraphics[width=1\textwidth]{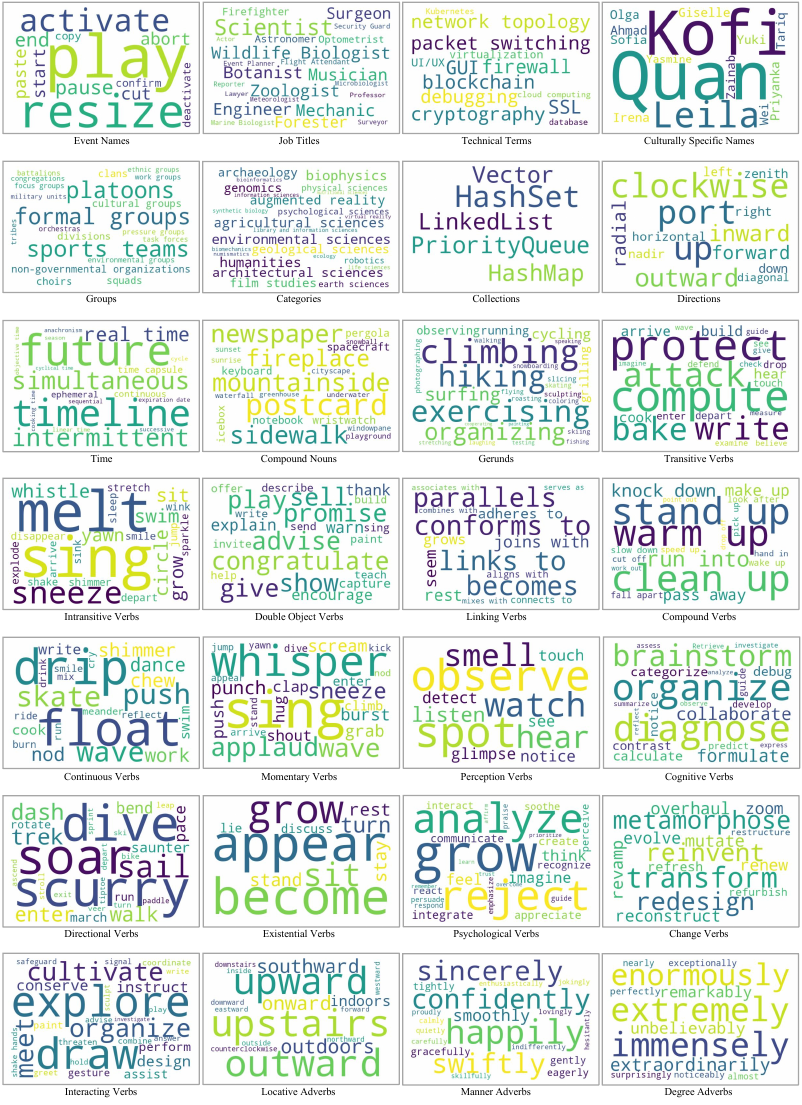}
  \caption{\label{fig_keywords2}\textbf{Example of keywords}. For each level-2 tag, we displayed up to 15 keywords through the word cloud, with random sampling and text size.}
\end{figure}

\begin{figure}[H]
  \centering
  \includegraphics[width=1\textwidth]{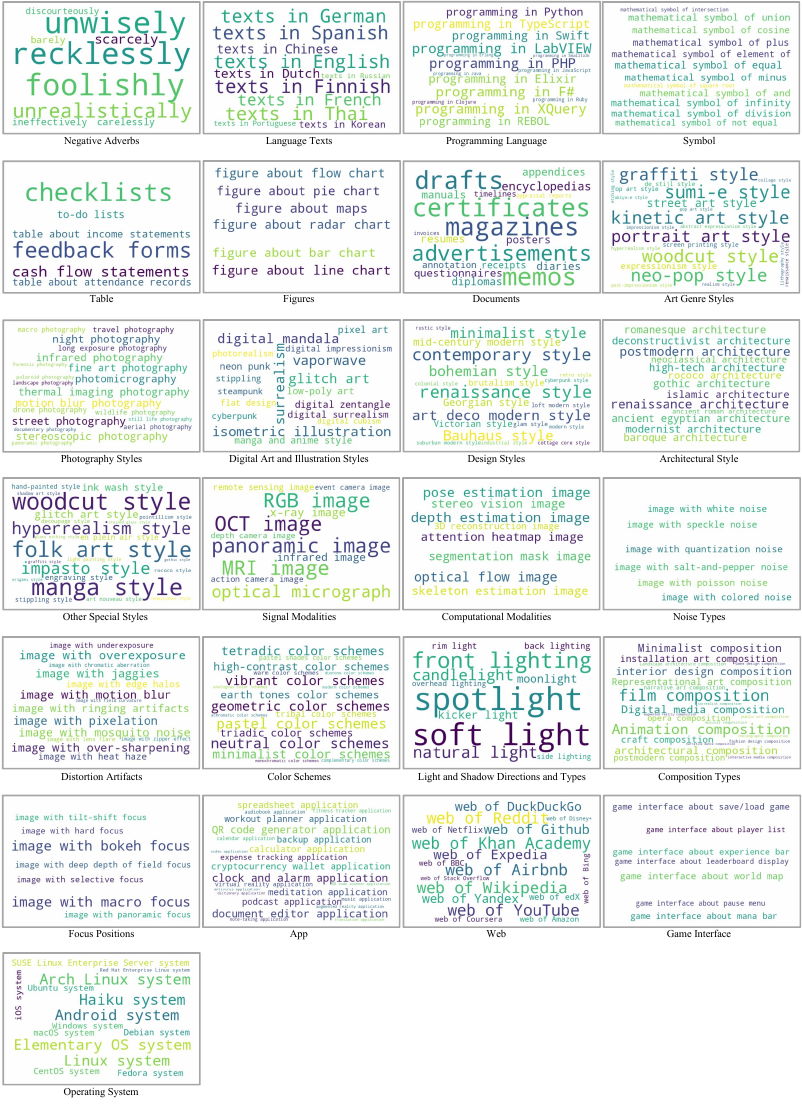}
  \caption{\label{fig_keywords3}\textbf{Example of keywords}. For each level-2 tag, we displayed up to 15 keywords through the word cloud, with random sampling and text size.}
\end{figure}

\vspace{-0.2cm}
\label{app_keywords}
We have provided keyword examples for each level-2 tag as shown in Fig. \ref{fig_keywords1}, Fig. \ref{fig_keywords2}, and Fig. \ref{fig_keywords3}. These keywords are used to bind prompts and generate questions. UniBench includes a wealth of detailed attributes to ensure diversity in evaluation. These tags include many novel attributes, such as time, emotions, celebrities, events, locations, actions, degrees, languages, symbols, programming, modalities, charts, figures, documents, UI, noise types, color schemes, lighting effects, composition, visual focus, and other new attributes.

\clearpage
\section{Details of UniBench Construction}
\label{app_unibench_gen}

\begin{figure}[H]
  \centering
    \vspace{-0.1cm}
  \includegraphics[width=1\textwidth]{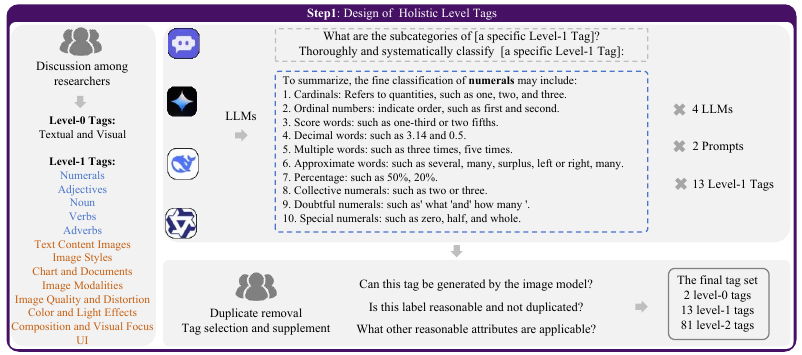}
  \vspace{-0.3cm}
  \caption{\label{step1}\textbf{Step 1 of UniBench Construction}. The gray box indicates the used prompts with answers marked by the blue box.}
\end{figure}

We have detailed the construction of UniBench in Sec. \ref{sec_31}. In this section, we highlight the prompts used for invoking LLMs and some details on processing the samples. In step 1 (Fig. \ref{step1}), we employed four LLMs to construct level-2 tags, including POE, Gemini-1.5-Pro \cite{team2024gemini}, Deepseek-R1-70B \cite{guo2025deepseek}, and Qwen2.5-72B \cite{yang2024qwen2} on webs. The prompts involved include ``What are the subcategories of [a specific Level-1 Tag]?'' and ``Thoroughly and systematically classify [a specific Level-1 Tag]:''. After collecting outputs from the LLMs, tags, and prompts, we performed deduplication and then manually selected appropriate level-2 tags. Our considerations included ``Can this tag be generated by the image model?'', ``Is this label reasonable and not duplicated?'', and ``What other reasonable attributes are applicable?''.

\vspace{-0.3cm}
\begin{figure}[H]
  \centering
  \includegraphics[width=1\textwidth]{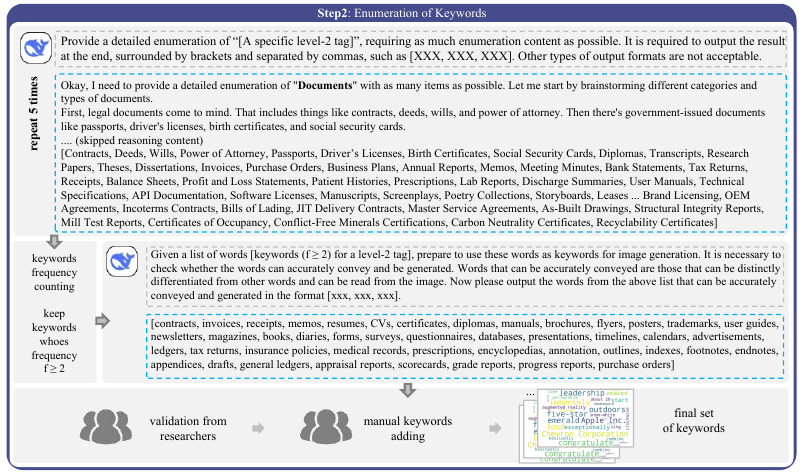}
  \vspace{-0.4cm}
  \caption{\label{step2}\textbf{Step 2 of UniBench Construction}. The gray box indicates the used prompts with answers marked by the blue box. The used LLM is Deepseek-R1-70B \cite{guo2025deepseek}.}
\end{figure}
\vspace{-0.2cm}

As shown in Fig. \ref{step2}, in the second step, we used Deepseek-R1-70B \cite{guo2025deepseek} to enumerate keywords for level-2 tags. The prompt used was ``Provide a detailed enumeration of "[A specific level-2 tag]", requiring as much enumeration content as possible. It is required to output the result at the end, surrounded by brackets and separated by commas, such as [xxx, xxx, xxx]. Other types of output formats are not acceptable.''. We required the model to run five times, retaining only keywords that appeared at least twice to minimize random errors. Then, we used the prompt to verify the keywords: ``Given a list of words [keywords (f $\geq$ 2) for a level-2 tag], prepare to use these words as keywords for image generation. It is necessary to check whether the words can accurately convey and be generated. Words that can be accurately conveyed are those that can be distinctly differentiated from other words and can be read from the image. Now please output the words from the above list that can be accurately conveyed and generated in the format [xxx, xxx, xxx].'', ensuring they were suitable for image generation. Finally, we conducted a manual review, adding appropriate keywords, and ultimately determined the final set of keywords.

\begin{figure}[H]
  \centering
  \includegraphics[width=1\textwidth]{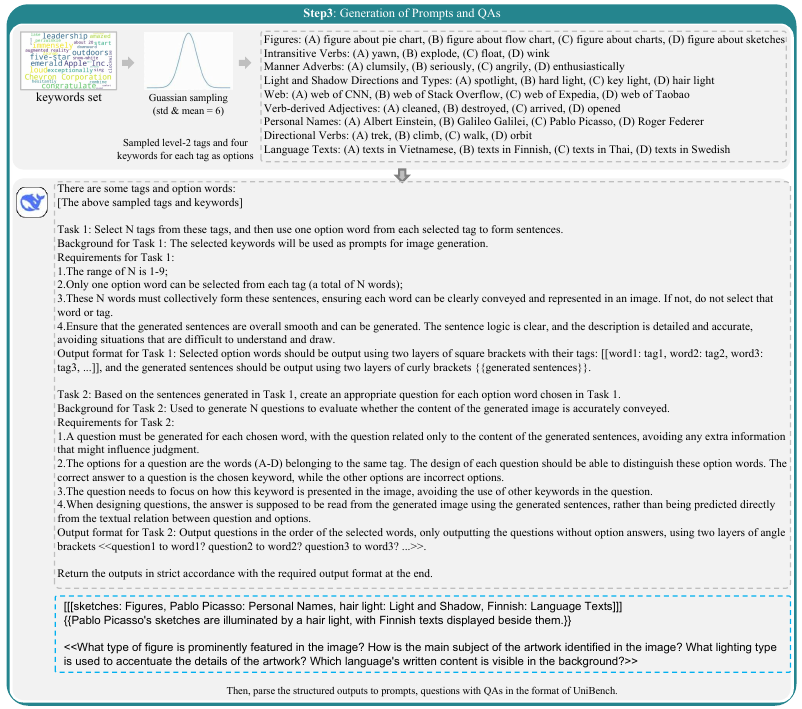}
    \vspace{-0.4cm}
  \caption{\label{step3}\textbf{Step 3 of UniBench Construction}. The gray box indicates the used prompts with answers marked by the blue box.}
  \vspace{-0.3cm}
\end{figure}

In the third step, we used a Gaussian sampling to randomly take N tags with four keywords from each tag as options, constructing the prompt input for the upper part of Fig. \ref{step3}. Combining this with the prompts for the lower part, we instructed Deepseek-R1-70B to choose keywords and generate sentences and questions. We defined two tasks in the prompts, clearly outlining the criteria. After the model produced structured outputs, we parsed them and obtained 2,000 initial cases.

\begin{figure}[H]
  \centering
  \vspace{-0.2cm}
  \includegraphics[width=1\textwidth]{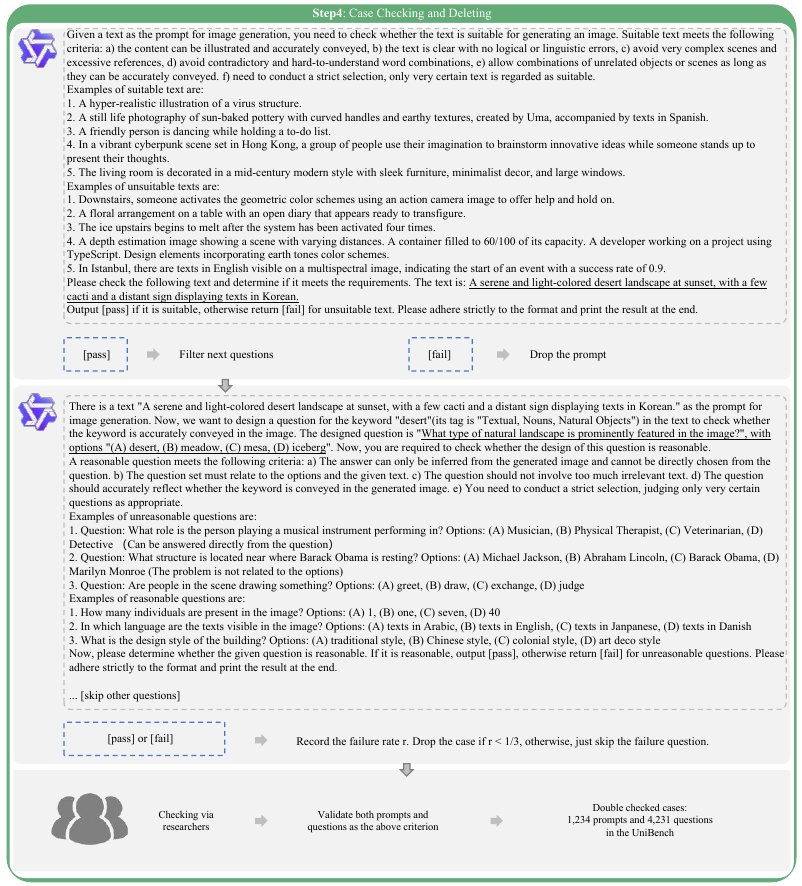}
  \vspace{-0.3cm}
  \caption{\label{step4}\textbf{Step 4 of UniBench Construction}. The gray box indicates the used prompts with answers marked by the blue box. The used LLM is QWQ-32b \cite{qwq32b}.}
\end{figure}

To further ensure the quality of the benchmark, we conducted validation as shown in Fig. \ref{step4}. First, we introduced another LLM, QWQ-32B \cite{qwq32b}, to verify the initial prompts generated by Deepseek-R1-70B \cite{guo2025deepseek}. Our requirement was ``Suitable text meets the following
criteria: a) the content can be illustrated and accurately conveyed, b) the text is clear with no logical or linguistic errors, c) avoid very complex scenes and excessive references, d) avoid contradictory and hard-to-understand word combinations, e) allow combinations of unrelated objects or scenes as long as they can be accurately conveyed. f) need to conduct a strict selection, only very certain text is regarded as suitable.''. We also provided five positive and five negative examples as in-context examples to assist the LLM in making judgments. Prompts for visual generation that did not meet the requirements were discarded, while those that did proceeded to the next step of question validation. We also used QWQ-32B to check quesitons with requirements: ``Now, you are required to check whether the design of this question is reasonable.
A reasonable question meets the following criteria: a) The answer can only be inferred from the generated image and cannot be directly chosen from the question. b) The question set must relate to the options and the given text. c) The question should not involve too much irrelevant text. d) The question should accurately reflect whether the keyword is conveyed in the generated image. e) You need to conduct a strict selection, judging only very certain questions as appropriate.''. We also supplied three positive and three negative examples as in-context examples. Finally, we calculated the ratio r of failed questions in each case. If r was less than 1/3, we deleted the unsuitable questions; otherwise, we considered the effective keywords too few and skipped the entire case. Subsequently, we conducted a manual review using the same criteria to ensure the quality of prompts and questions. Ultimately, we confirmed 1,234 prompts and 4,231 QAs.

\section{Human Evaluation Cases}
\label{app_humaneval_case}
\begin{figure}[H]
  \centering
  \vspace{-0.2cm}
  \includegraphics[width=1\textwidth]{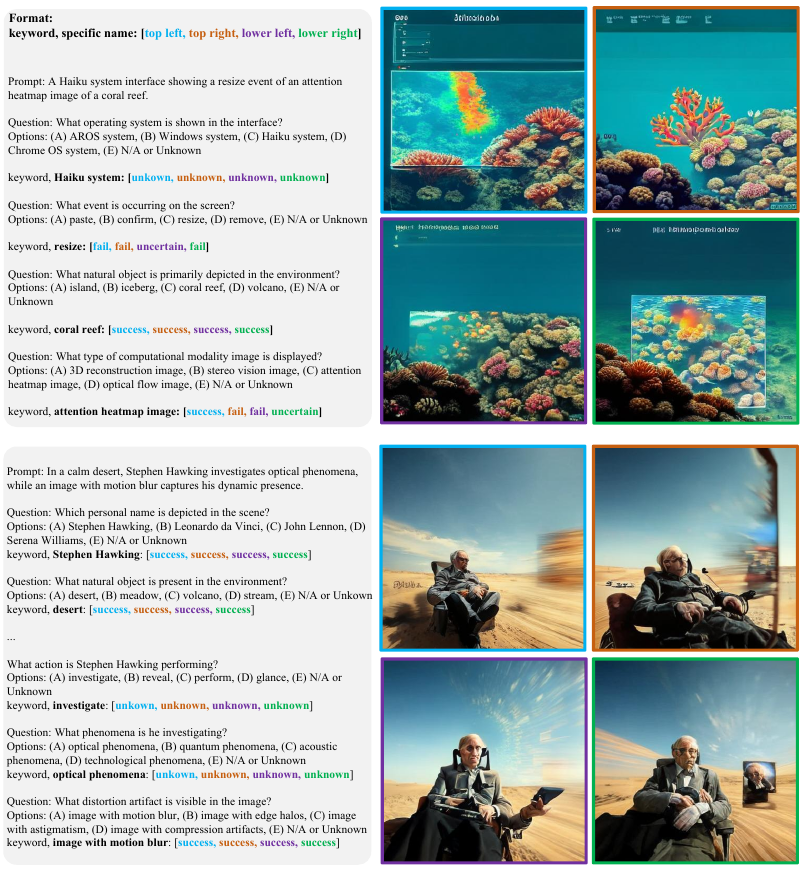}
  \vspace{-0.4cm}
  \caption{\label{vis_anno}\textbf{Visualization of Annotations}. The prompts, questions, and labeled results are listed on the left, with generated images on the right. The label colors correspond to the image with the same color board. The uncertain label is related to ambiguous generation, while the unknown means unable to judge. These two labels are excluded from the evaluation to ensure reliability. These results are from annotator 1 (undergraduate background) on Janus-Pro-7B \cite{chen2025janus}.}
\end{figure}

In Sec. \ref{sec_33}, we have introduced the criterion of human evaluation: ``The annotators were asked to label whether the keywords were expressed in the generated images for each question. Given the complexity of labels and the subjectivity in generation, we provided four labels: (1) generation failure, (2) between success and failure, (3) successful generation, and (0) lacking knowledge to judge''. In this section, we provide some visualized annotation results to help readers understand our annotation process. Fig. \ref{vis_anno} shows annotation examples from annotator 1 (Undergraduate background) on Janus-Pro-7B \cite{chen2025janus}. We list the prompt and quetions on the left for reference with corresponding generated images on the right of this figure. The annotations are colored the same as the image borders, including failure (1), uncertain (2), success (3), and unknown (0). These two examples are complex, where annotator 1 annotated four types of labels in the first case, and marked more unknown labels for the second one.

\section{Human Evaluation Analysis}
\label{app_humaneval_ana}
\begin{figure}[H]
  \centering
  \vspace{-0.1cm}
  \includegraphics[width=1\textwidth]{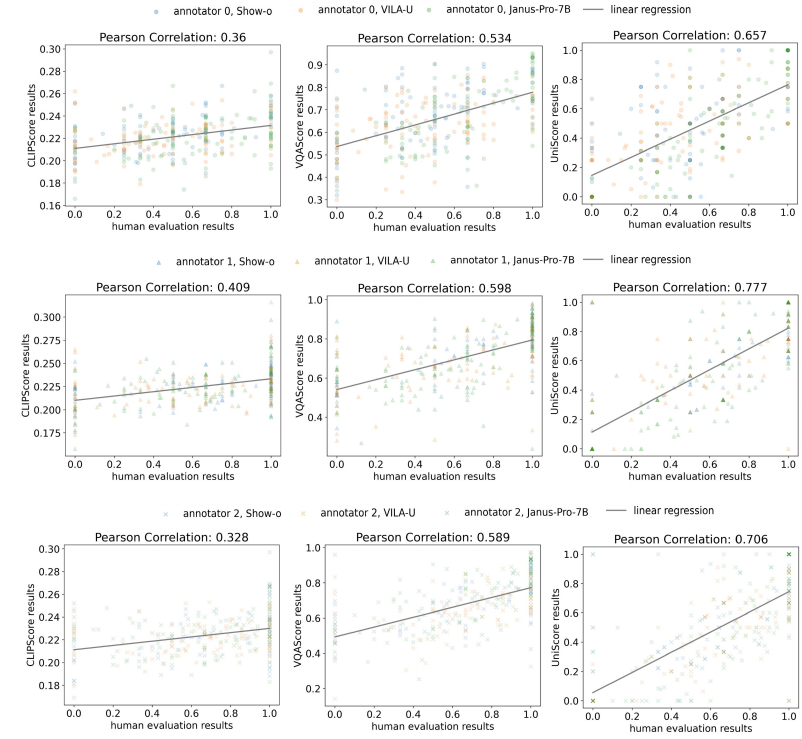}
  \vspace{-0.3cm}
  \caption{\label{fig_human_eval_annot}.\textbf{Human Study in the Annotator Aspect}. The first, second, and third rows indicate human studies from annotator 0 (PhD), 1 (undergraduate), and 2 (master), respectively. The Pearson Correlation \cite{cohen2009pearson} is a normalized covariance to measure the alignment between auto evaluation metrics (CLIPScore \cite{hessel2021clipscore}, VQAScore \cite{li2024evaluating}, our UniScore) and human evaluations.}
  \vspace{-0.4cm}
\end{figure}

\begin{figure}[H]
  \centering
  \vspace{-0.1cm}
  \includegraphics[width=1\textwidth]{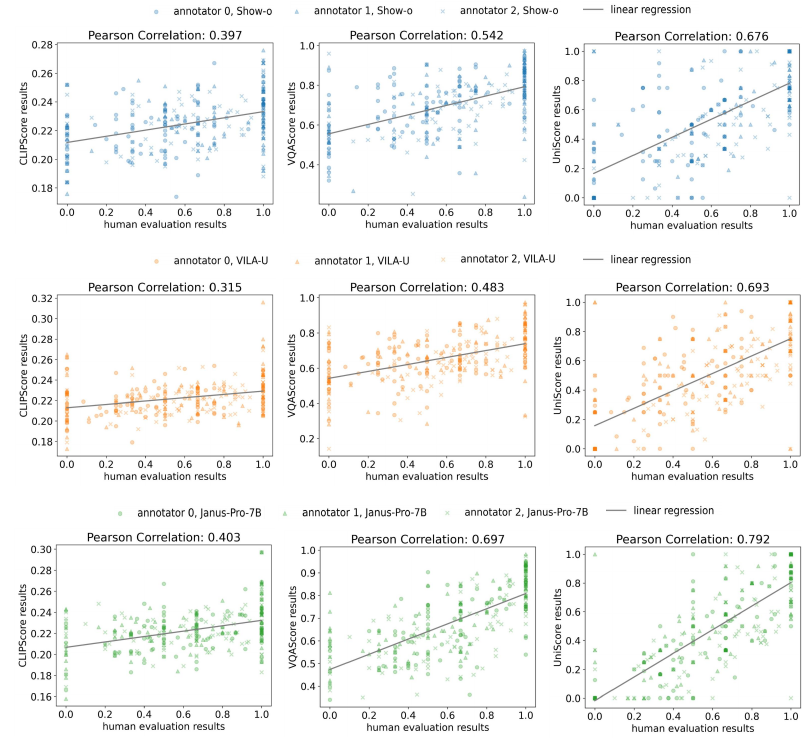}
  \vspace{-0.3cm}
  \caption{\label{fig_human_eval_models}\textbf{Human Study in the Model Aspect}. The first, second, and third rows indicate human studies conducted on Show-o \cite{xie2024show}, VILA-U \cite{wu2024vila}, and Janus-Pro-7B \cite{chen2025janus}, respectively. The Pearson Correlation \cite{cohen2009pearson} is a normalized covariance to measure the alignment between auto evaluation metrics (CLIPScore \cite{hessel2021clipscore}, VQAScore \cite{li2024evaluating}, our UniScore) and human evaluations.}
\end{figure}

\vspace{-0.1cm}
In addition to the overall human study in Fig. \ref{fig_human_eval}, we conducted human studies in the aspect of varied annotators (Fig. \ref{fig_human_eval_annot}) and different models (Fig. \ref{fig_human_eval_models}) using the same 300 random cases from UniBench. Fig. \ref{fig_human_eval_annot} shows the correlation results from different annotators. The first row presents the results from annotator 0 (PhD background), with the lowest overall scores, indicating a stricter evaluation criterion and a lower tendency to label responses as unknown. The second row comes from annotator 1 (undergraduate background), who performed the best overall, with a correlation of 0.777 between UniScore and human evaluations. The results from annotator 2 (master background) are between other annotators. From the three different annotators, we found that UniScore is consistently higher than the recent VQAScore \cite{li2024evaluating} and significantly exceeds the commonly used instruction-following metric, CLIPScore \cite{hessel2021clipscore}.

Fig. \ref{fig_human_eval_models} compares the correlation of various metrics with human evaluations across different models. The results in the first row come from Show-o \cite{xie2024show}, marked in blue; the second row is from VILA-U \cite{wu2024vila}, in orange; and the third row presents the results from Janus-Pro-7B \cite{chen2025janus}. Consistent with the overall results and the annotator aspect, the proposed UniScore shows superior alignment compared to existing metrics across different models, surpassing VQAScore \cite{li2024evaluating} by 0.134, 0.21, and 0.095, respectively. This is attributed to UniScore providing more options and relevant prompts, which can reduce random errors beyond binary options while offering correlational information from the prompt and options related to keywords, thereby avoiding ambiguity. In contrast, CLIPScore is based solely on keyword similarity, while VQAScore relies on prompt templates. The experiments indicate that UniScore is a robust instruction-following metric that aligns well with human perception.

\vspace{-0.3cm}
\section{Comparision with Larger Extra Model}
\label{app_larger_model}
\vspace{-0.1cm}
In human study and visual generation evaluation, we introduced an extra model, Qwen2.5-VL-7B \cite{bai2025qwen2}. In this analysis, we also compared it to the larger Qwen2.5-VL-72B \cite{bai2025qwen2} to explore the relationship between model scale and human alignment. It should be noted that both models outperformed the closed-source GPT-4v on MMMU \cite{yue2024mmmu}, while the 72B model's performance is closer to GPT-4o. We avoided using closed-source models to mitigate the usage costs for followers, and Qwen2.5-VL-7B is currently the strongest 7B model on MMMU.

\begin{figure}[H]
  \centering
    \vspace{-0.2cm}
  \includegraphics[width=1\textwidth]{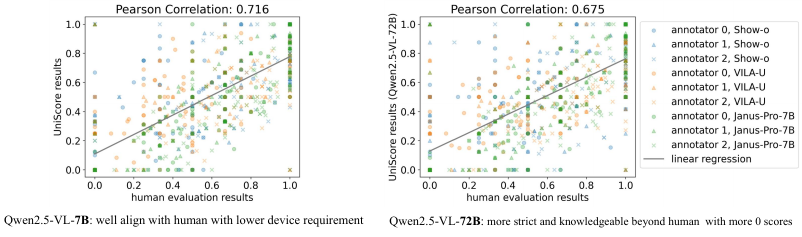}
    \vspace{-0.4cm}
  \caption{\label{vs_72b_model}\textbf{Comparision with Larger Model in Alignment}. The alignment degrees in the human study (measured by Pearson correlation) are close between the 7B and 72B models. Since Qwen2.5-VL-72B is stricter (lower UniScore when evaluating the same model) than the 7B model, its alignment degree is slightly lower than the 7B model. Qwen2.5-VL-7B \cite{bai2025qwen2} is the best 7B model on MMMU \cite{bai2025qwen2} currently, and saves API cost compared with close-set models, as well as GPU memory compared with larger models.}
    \vspace{-0.3cm}
\end{figure}

In Fig. \ref{vs_72b_model}, we compared the alignment of Qwen2.5-VL-7B and Qwen2.5-VL-72B with humans on UniBench (using the same models and sampled cases as human evaluations). The results show that the performance of both models is similar, with the 7B model having slightly better alignment. This is mainly because the criteria for the 72B model are stricter; for example, in the evaluation of Janus-Pro-7B, the UniScore for the 72B model is 0.388, while for the 7B model, it is 0.402. Compared to the 72B model, human expectations for the success of the generation are not as strict, which leads to better alignment for Qwen2.5-VL-7B. Additionally, the 7B model has lower memory usage, significantly reducing the hardware requirements for the visual generation part. Note that the Unified model does not require additional understanding of the model; most models can run on a single 24GB memory GPU, while visual generation model evaluations generally require two 24GB GPUs with an extra understanding model.

\vspace{-0.3cm}
\section{UniEval for Task-specific Evaluations.}
\label{app_specific_eval}
\vspace{-0.2cm}


In this section, we independently evaluate the generation and understanding abilities of unified models. This helps researchers to analyze the model strengths and weaknesses on specific tasks, providing more insights besides the overall results. It also proves the wide applicability of UniEval.

\begin{table}[htbp]
  \caption{\label{tab_extra_model}\textbf{Evaluation for the Visual Generation Part of Unified Models}. We apply the Qwen2.5-VL-7B \cite{bai2025qwen2} to evaluate the task-specific visual generation performances. By comparing unified models with visual generation models using the same model, we find new insights discussed in Sec. \ref{sec_insights}.}
  \setlength{\tabcolsep}{3.2pt}
  \scriptsize
  \centering
  \begin{tabular}{ccccccccccccccc}
    \hline
    \textbf{Model} & \textbf{Num} & \textbf{Adj} & \textbf{Noun} & \textbf{Verb}  &  \textbf{Adv} &  \textbf{Text} &  \textbf{Doc}  &  \textbf{Sty} &  \textbf{Moda} &  \textbf{Qual} & \textbf{Effe} &  \textbf{Comp} &  \textbf{UI} & \textbf{UniScore}$\uparrow$\\    \hline
    VARGPT \cite{zhuang2025vargpt} & 0.036 &0.168 &0.114 &0.074 &0.066 &0.010 &0.035 &0.114 &0.116 &0.120 &0.169 &0.164 &0.032 & 0.094 \\
    JanusFlow-1.3B \cite{ma2024janusflow} & 0.141 &0.483 &0.354 &0.267 &0.234 &0.104 &0.186 &0.608 &0.328 &0.340 &0.539 &0.480 &0.146 & 0.324 \\
    Janus-Pro-1B \cite{chen2025janus} & 0.126 &0.478 &0.334 &0.243 &0.237 &0.135 &0.213 &0.611 &0.442 &0.340 &0.575 &0.486 &0.085 & 0.331 \\
    VILA-U \cite{wu2024vila} & 0.148 &0.548 &0.367 &0.337 &0.258 &0.036 &0.208 &0.660 &0.392 &0.272 &0.609 &0.608 &0.159 & 0.354 \\
    TokenFlow \cite{qu2024tokenflow} & 0.178 &0.529 &0.405 &0.292 &0.284 &0.136 &0.249 &0.660 &0.401 &0.242 &0.574 &0.589 &0.224 & 0.366 \\
    Janus-1.3B \cite{wu2024janus} & 0.174 &0.523 &0.387 &0.312 &0.306 &0.120 &0.262 &0.650 &0.387 &0.418 &0.618 &0.541 &0.148 & 0.373 \\
    UniToken-II \cite{jiao2025unitoken} & 0.172 &0.538 &0.434 &0.361 &0.266 &0.109 &0.301 &0.601 &0.366 &0.348 &0.626 &0.552 &0.215 & 0.376 \\
    Show-o-Turbo \cite{xu2025show} & 0.180 &0.569 &0.452 &0.344 &0.268 &0.103 &0.271 &0.660 &0.390 &0.351 &0.662 &0.529 &0.197 & 0.383 \\
    Janus-Pro-7B \cite{chen2025janus} & 0.194 &0.584 &0.464 &0.358 &0.267 &0.167 &0.391 &0.682 &0.462 &0.322 &0.637 &0.562 &0.230 & 0.409 \\
    Show-o \cite{xie2024show} & 0.202 &0.611 &0.483 &0.397 &0.290 &0.126 &0.419 &0.659 &0.453 &0.355 &0.653 &0.469 &0.280 & 0.415 \\
    \hline
  \end{tabular}
    \vspace{-0.2cm}
\end{table}

For the evaluation of visual generation, we introduced the same understanding model Qwen2.5-VL-7B \cite{bai2025qwen2} as visual generation models in Table \ref{tab_gen}, focusing on comparing its visual generation part in a fair setting. It can be seen that Show-o \cite{xie2024show} achieves the highest pure generation capability at 0.415, while the unified model scores 0.367 in Tab. \ref{tab_uni}, indicating good generation quality but limited in understanding. Show-o ranks first in Table \ref{tab_extra_model} mainly because of a relatively high resolution (512), while other unified models typically use resolutions of 224 or 336. However, compared to the pure generation models in Table \ref{tab_gen}, the understanding models overall demonstrate weaker generation capabilities owing to lower resolution. For instance, among ten models, only two unified models have a UniScore exceeding 0.4 under understanding models, whereas seven pure generation models exceed 0.4. The ones that do not exceed 0.4 are primarily those with a resolution of 512 (models with results higher than 0.4 are all 1024 in resolution). This indicates that high resolution is very important in complex generation scenarios, and the unified model is lacking in this regard. Another reason is that the unified model must balance generation and understanding, which may lead to shortcomings in generation performance. Through the visual comparisons in Appendix \ref{app_cases}, we find that there is still room for improvement in its generation quality. In another aspect, we believe unified models are promising. At the same resolution, 512, the unified Show-o still outperforms generation-only PixArt-$\alpha$ \cite{chen2023pixart}, suggesting the potential of the unified model. 

\begin{table}[htbp]
  \caption{\label{tab_und}\textbf{Evaluation for the Understanding Part of Unified Models}. We evaluate the understanding ability of the unified model itself by the difference between the overall results (Table \ref{tab_uni}) and the understanding results in Table \ref{tab_extra_model}. The results are $\Delta = Uni.-Gen.$, measuring the understanding ability difference between the unified model and the extra model. A positive value indicates that the unified model outperforms the extra model (Qwen2.5-VL-7B \cite{bai2025qwen2}) in understanding generated images, highlighting understanding as its strength. A negative value indicates that the understanding of the unified model is weaker than that of the extra model, highlighting understanding as its weakness. The best understanding result is marked in green, while the lowest value is marked in red.}
  \setlength{\tabcolsep}{2.5pt}
  \scriptsize
  \centering
  \begin{tabular}{ccccccccccccccc}
    \hline
    \textbf{Model} & \textbf{Num} & \textbf{Adj} & \textbf{Noun} & \textbf{Verb}  &  \textbf{Adv} &  \textbf{Text} &  \textbf{Doc}  &  \textbf{Sty} &  \textbf{Moda} &  \textbf{Qual} & \textbf{Effe} &  \textbf{Comp} &  \textbf{UI} & \textbf{UniScore}$\uparrow$\\    \hline
VARGPT \cite{zhuang2025vargpt} & 0.061 & 0.158 & 0.17 & 0.214 & 0.144 & \textcolor{red}{0.039} & 0.069 & 0.113 & \textcolor{green}{0.111} & \textcolor{green}{0.215} & \textcolor{green}{0.072} & -0.009 & 0.077 & 0.11\\
TokenFlow \cite{qu2024tokenflow} & \textcolor{red}{-0.085} & -0.007 & -0.017 & 0.038 & -0.009 & 0.021 & -0.026 & -0.06 & -0.049 & \textcolor{red}{-0.079} & -0.039 & -0.072 & \textcolor{red}{-0.061} & -0.034\\
Show-o-Turbo \cite{xu2025show} & 0.07 & \textcolor{red}{-0.267} & \textcolor{red}{-0.099} & -0.07 & -0.012 & 0.278 & 0.06 & \textcolor{red}{-0.274} & \textcolor{red}{-0.059} & 0.195 & -0.268 & \textcolor{red}{-0.169} & 0.201 & -0.032\\
Show-o \cite{xie2024show} & 0.048 & -0.249 & -0.061 & \textcolor{red}{-0.081} & -0.005 & 0.255 & \textcolor{red}{-0.061} & -0.269 & -0.107 & 0.117 & \textcolor{red}{-0.221} & -0.109 & 0.118 & \textcolor{red}{-0.048}\\
Janus-Pro-1B \cite{chen2025janus} & 0.06 & 0.026 & 0.109 & 0.17 & 0.133 & \textcolor{red}{0.039} & 0.02 & -0.075 & 0.061 & 0.056 & -0.178 & -0.136 & 0.216 & 0.039\\
Janus-1.3B \cite{wu2024janus} & 0.028 & -0.039 & 0.11 & 0.072 & \textcolor{red}{-0.022} & 0.126 & 0.057 & -0.009 & -0.006 & -0.01 & -0.142 & -0.118 & \textcolor{green}{0.301} & 0.027\\
VILA-U \cite{wu2024vila} & 0.083 & 0.056 & 0.191 & 0.212 & 0.139 & 0.218 & \textcolor{green}{0.168} & 0.044 & 0.175 & 0.09 & -0.017 & -0.155 & 0.126 & 0.102\\
UniToken-II \cite{jiao2025unitoken} & 0.177 & 0.099 & 0.19 & 0.204 & 0.12 & 0.168 & 0.129 & 0.068 & 0.227 & -0.019 & -0.058 & \textcolor{green}{0.037} & 0.165 & 0.116\\
JanusFlow-1.3B \cite{ma2024janusflow} & \textcolor{green}{0.183} & 0.125 & \textcolor{green}{0.234} & 0.261 & 0.189 & \textcolor{green}{0.296} & \textcolor{green}{0.168} & 0.098 & 0.317 & 0.181 & 0.046 & 0.016 & 0.28 & \textcolor{green}{0.184}\\
Janus-Pro-7B \cite{chen2025janus} & 0.162 & \textcolor{green}{0.132} & 0.202 & \textcolor{green}{0.263} & \textcolor{green}{0.242} & 0.289 & 0.086 & 0.095 & 0.21 & 0.22 & 0.018 & -0.035 & 0.229 & 0.163\\
\hline
  \end{tabular}
    \vspace{-0.2cm}
\end{table}


For evaluating the model's understanding ability, using an extra generation model is not appropriate because existing models perform inadequately. Our approach is to compare the overall results of the unified model with those of the visual generation model. This ensures that the generation model is the same, allowing a fair and reasonable comparison between the understanding ability of the unified model and that of the extra understanding model. As shown in Table \ref{tab_und}, we compare the unified overall results with the results based on the extra model, calculating the difference $\Delta = Uni.- Gen.$, which measures the difference in understanding ability between the unified model and the extra model. A positive value indicates that the unified model outperforms the extra model (Qwen2.5-VL-7B \cite{bai2025qwen2}) in understanding generated images, highlighting understanding as its strength. A negative value indicates that the unified model’s understanding is weaker than that of the extra model, highlighting understanding as its weakness. In the table, the best results for understanding are marked in green, while the worst results are marked in red. We observe that Janus-Flow-1.3B \cite{ma2024janusflow} and Janus-Pro-7B \cite{chen2025janus} have the best understanding of generated images. In Sec. \ref{sec_insights}, we refer to this ability as self-consistency, meaning the model’s capability to accurately understand the images it generates itself. The models with the largest understanding bias are Show-o \cite{xie2024show}, Show-o-Turbo \cite{xu2025show}, and TokenFlow \cite{qu2024tokenflow}. The reasons for this are also analyzed in Table \ref{tab_insights}, mainly attributed to bias in the understanding model’s responses. This reflects that the understanding ability of these models needs improvement. Additionally, although VARGPT \cite{zhuang2025vargpt} has three green marks, this model often refuses to generate images, resulting in an overall low score with limited reference value. Overall, the scores of unified models are higher than those of the extra models, with most values being positive. This indicates that unified models generally have better understanding abilities for generated images than understanding-only models, demonstrating the unique value of unified models.

\vspace{-0.3cm}
\section{Comparison with T2I Benchmarks.}
\label{app_comp_t2i}
\vspace{-0.2cm}
We compared the difficulty, discriminability, and diversity of across text-to-image benchmarks in Fig. \ref{fig_vs_bench}. In this section, we reports the specific data involved in Table \ref{tab_specific_data}. Due to differet evaluated models reported by different benchmarks, we selected five commonly used models for comparison, including SDV1 \cite{rombach2022high}, SDV2 \cite{rombach2022high}, PixArt-$\alpha$ \cite{chen2023pixart}, SDXL \cite{podell2023sdxl}, and DALL-E3 \cite{betker2023improving}. Among these, only T2I-CompBench++ \cite{huang2025t2i} uses SDv1.4 and SDv2.0; the others are SDv1.5 and SDv1.4. We applied min-max normalization to the quantity for a normalized overall result, which was counted at the level-1 tags, as some benchmarks do not contain fine-grained labels. Among them, T2I-CompBench++ is evaluated by multiple models and does not have an overall metric; we report its complexity metric evaluated by GPT-4v \cite{achiam2023gpt}. Based on this data, we quantified the comparisons in various aspects. Fig. \ref{fig_vs_bench} indicates that our UniBench significantly outperforms existing benchmarks, with an overall value of 0.961, notably exceeding the second-best GenEval \cite{ghosh2023geneval} of 0.466.

\begin{table}[H]
  \vspace{-0.2cm}
  \caption{\label{tab_specific_data}\textbf{Specific Data in Benchmark Comparision}. Different benchmarks use diverse models, thus, we pick these five common models for fair comparison. The category number is min-max normalized to count the normalized average.}
  \setlength{\tabcolsep}{2pt}
  \label{tab_vs_gen_bench}
  \scriptsize
  \centering
  \begin{tabular}{cccccc}
    \hline
    \textbf{Models / Aspects} & \textbf{GenEval} \cite{ghosh2023geneval} & \textbf{DPG-bench} \cite{hu2024ella} & \textbf{T2I-CompBench++} \cite{huang2025t2i} & \textbf{ConceptMix} \cite{wu2024conceptmix} & \textbf{UniBench (ours)} \\
    \hline
    SDv1 \cite{rombach2022high} & 0.43 & 0.6318 & 0.6453 & 0.52 & 0.33 \\
    SDv2 \cite{rombach2022high} & 0.5 & 0.6809 & 0.6483 & 0.52 & 0.355 \\
    PixArt-$\alpha$ \cite{chen2023pixart} & 0.48 & 0.7111 & 0.7223 & 0.66 & 0.379 \\
    SDXL \cite{podell2023sdxl} & 0.55 & 0.7465 & 0.717 & 0.69 & 0.404 \\
    DALL-E3 \cite{betker2023improving} & 0.67 & 0.835 & 0.8653 & 0.83 & 0.526 \\
    \hline
    Min Error Rate (Difficulty) & 0.33 & 0.165 & 0.135 & 0.17 & 0.474 \\
    Coefficient of variation (Discriminability) & 0.155 & 0.095 & 0.111 & 0.181 & 0.171 \\
    Category Number (Diversity) & 6 & 5 & 6 & 7 & 13 \\
    Normalized Average (Overall) & 0.466 & 0.03 & 0.104 & 0.451 & 0.961 \\
    \hline
  \end{tabular}
    \vspace{-0.2cm}
\end{table}

\vspace{-0.5cm}
\section{Limitation and Broader Impacts}
\label{app_limitaion}
\vspace{-0.2cm}


Although UniEval, as the first evaluation framework designed for unified models, addresses many limitations of existing task-specific benchmarks, limitations still objectively exist. First, our evaluation framework emphasizes instruction-following and does not include image quality assessment following most text-to-image benchmarks \cite{huang2025t2i,wu2024conceptmix,ghosh2023geneval,hu2024ella}. If we add evaluate metrics like FID \cite{heusel2017gans}, additional images and models would need to be introduced, which goes against our motivation. Second, UniEval emphasizes overall evaluation, only partially achieving individual assessment. Although our UniBench, combined with extra models \cite{bai2025qwen2}, shows significant advantages in evaluation of visual generation compared to conventional text-to-image benchmarks \cite{huang2025t2i,wu2024conceptmix,ghosh2023geneval,hu2024ella}, the current approach to analyzing understanding ability is limited to comparing overall results and visual generation results as discussed in Appendix \ref{app_specific_eval}. We attempted to fix the visual generation model to generate a fixed dataset for evaluating understanding ability, but due to the limitations of existing model generation capabilities, we cannot directly evaluate understanding ability without human efforts to select the correct generated images. Third, ensuring the quality of the benchmark still incurs human effort. Although we do not need annotators to label ground truth, the LLMs sometimes generate overly complex and unsuitable prompts. The quality of questions can occasionally be poor, such as when answers from options appear in the questions. This necessitates the introduction of a certain level of manual checking costs to ensure quality (still far less than directly annotating answers).


Opportunities and challenges coexist. As the first unified evaluation framework, there is still much room for improvement. For example, it is possible to incorporate more diverse content, such as image quality assessments, with minimal additional resources. Alternatively, followers could select specific generation models and include manual screening to create a benchmark for generated images, accommodating both overall and task-specific evaluations. Additionally, enhancing the quality and quantity of synthesis could enable more detailed evaluations, such as requiring specific textual content beyond just language type. We believe that unified evaluations will be a great pathway to achieve simplified, comprehensive, convenient, and high-quality evaluations. This approach also has strong potential to generalize to future unified models encompassing video, audio, and other capabilities, establishing a new standard for multimodal model evaluation, as well as inspiring the development of more powerful models and applications.

\section{Case Study}
\label{app_cases}

\begin{figure}[H]
  \centering
  \vspace{-0.2cm}
  \includegraphics[width=1\textwidth]{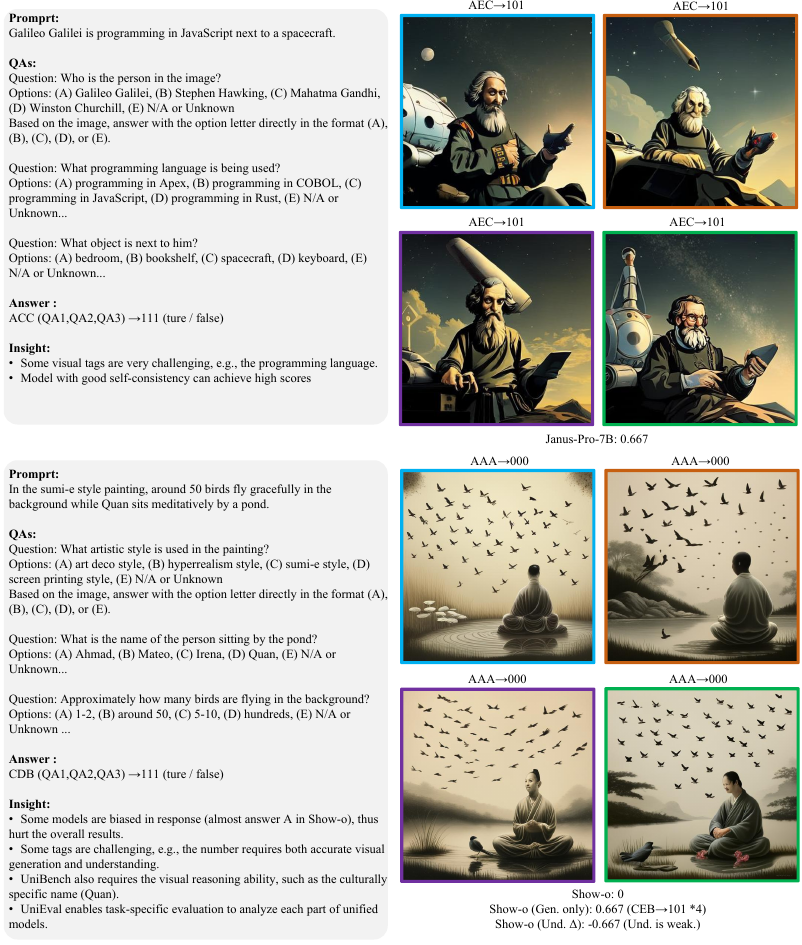}
  \vspace{-0.3cm}
  \caption{\label{fig_cases_model1}. \textbf{Insight Analysis From Cases of Unified Models}. Some visual tags are very challenging, e.g., the programming language, some attributes also require the visual reasoning ability, such as the culturally specific name (Quan). We find model with good self-consistency can achieve high scores. While some models are biased in response (almost answer A in Show-o), thus, hurt the overall results. Besides, UniEval enables task-specific evaluation to analyze each part of unified models as in the case of Show-o \cite{xie2024show}. The case information is shown on the left, including prompt, questions, options, answers, and insights. The generated images are depicted on the right, with corresponding model predictions on the image top. The model name is listed on the bottom right, with the UniScore of this case. The Gen. only score and Und. score is calculated as introduced in Appendix \ref{app_specific_eval}.}
\end{figure}

\begin{figure}[H]
  \centering
  \includegraphics[width=1\textwidth]{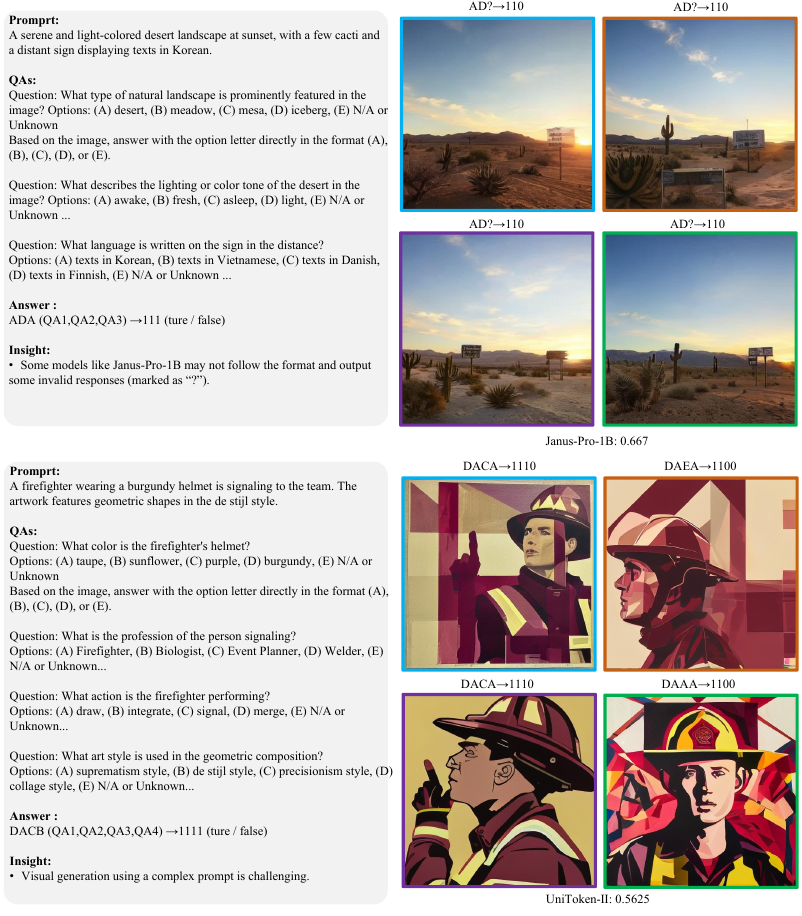}
  \vspace{-0.3cm}
  \caption{\label{fig_cases_model2}. \textbf{Insight Analysis From Cases of Unified Models}. Some models like Janus-Pro-1B \cite{chen2025janus} may not follow the format and output some invalid responses (marked as “?”). Besides, visual generation using a complex prompt is challenging. The case information is shown on the left, including prompt, questions, options, answers, and insights. The generated images are depicted on the right, with corresponding model predictions on the image top (``?'' indicates an invalid response out of A-E). The model name is listed on the bottom right, with the UniScore of this case. }
\end{figure}

We visualized example cases of UniEval in Fig. \ref{fig_cases_model1} and Fig. \ref{fig_cases_model2} with analysis of insights. On the left side of the image, we showcase the visual generation prompts, multimodal understanding questions, options, and inference prompts. In the bottom, we present sample answers and evaluation results for each question. Finally, we emphasize the insights corresponding to this case. On the right side, there are four images generated by the model; the top of the images corresponds to the model's outputs, and whether they are correct. At the bottom right, we provide specific model names and case-level UniScore.

From Fig. \ref{fig_cases_model1}, we find some visual tags are very challenging, e.g., the programming language. Besides, a model with good self-consistency can achieve high scores. The second case in this figure shows that some models are biased in response (almost answer A in Show-o \cite{xie2024show}). Thus, hurt the overall results. Besides, tags like number require accurate both visual generation and understanding, which is challenging. The case also tells us that UniBench requires the visual reasoning ability, such as the culturally specific name (Quan). Moreover, UniEval enables task-specific evaluation to analyze each part of unified models, where the generation ability of Show-o is good, but the understanding is wrong. From Fig. \ref{fig_cases_model2}, we find some models like Janus-Pro-1B\cite{chen2025janus} may not follow the format and output some invalid responses (the ``?'' on the image top). The second case shows that visual generation using a complex prompt is challenging. These examples provide templates for the case study, which help researchers to investigate the models and foster further improvements.

\begin{figure}[H]
  \centering
  \includegraphics[width=1\textwidth]{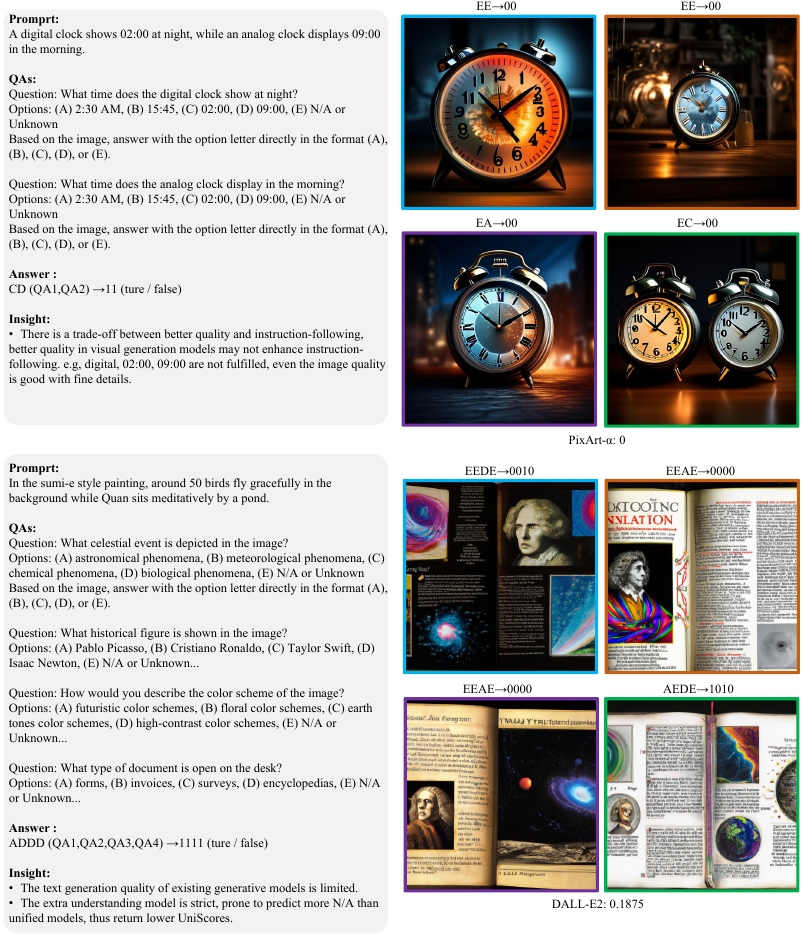}
  \vspace{-0.3cm}
  \caption{\label{fig_case_gen1}. \textbf{Insight Analysis From Cases of Visual Generation Models}. There is a trade-off between better quality and instruction-following; better quality in visual generation models may not enhance instruction-following. Besides, the text generation quality of existing generative models is limited. The extra understanding model is strict; prone to predict more N/A than unified models. The case information is shown on the left, including prompt, questions, options, answers, and insights. The generated images are depicted on the right, with corresponding model predictions on the image top. The model name is listed on the bottom right, with the UniScore of this case.}
\end{figure}

In Fig. \ref{fig_case_gen1}, we also analyze the cases of the visual generation models. We can see that the generation-only models are of high quality, especially with fewer flaws in the details. However, we find that there is a trade-off between image quality and instruction following. Better quality in visual generation models may not enhance instruction-following. For example, elements like digital, 02:00, and 09:00 are not accurately fulfilled, even though the image quality is good with fine details. These findings suggest that instruction-following is not a simple task, and further improvements are necessary, as measured by this challenging and diverse benchmark.


\begin{figure}[H]
  \centering
  \includegraphics[width=1\textwidth]{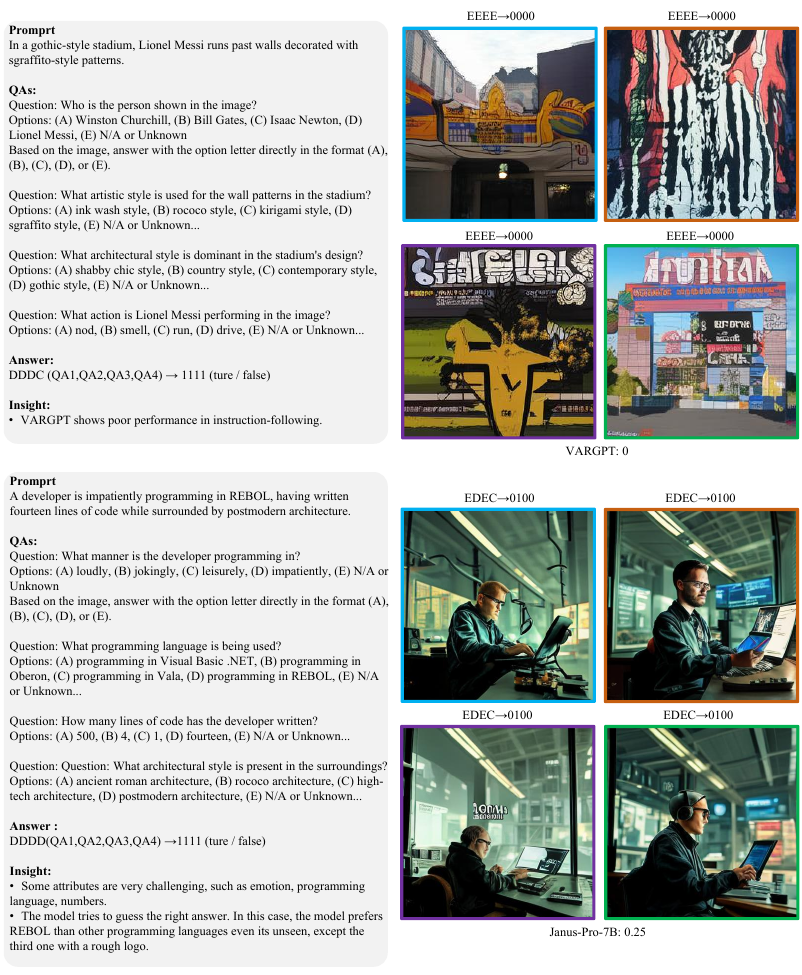}
  \vspace{-0.3cm}
  \caption{\label{fig_case3}. \textbf{Failure Case Analysis of Unified Models}. VARGPT \cite{zhuang2025vargpt} shows poor performance in instruction-following. Janus-Pro-7B \cite{chen2025janus} presents low performances for challenging attributes, such as emotion, programming language, and numbers. The case information is shown on the left, including prompt, questions, options, answers, and insights. The generated images are depicted on the right, with corresponding model predictions on the image top. The model name is listed on the bottom right, with the UniScore of this case.}
\end{figure}

We further showcased the challenging attributes and failure cases. As shown in Fig. \ref{fig_case3}, the instruction-following capability poses significant challenges for some models like VARGPT \cite{zhuang2025vargpt}, leading to difficulties in generating controllable images. We found that certain tags, such as emotion, programming
language, and numbers are particularly challenging. Additionally, we observed that unified models tend to guess answers, resulting in a higher UniScore compared to strict extra evaluation models. However, under the same additional models, their overall performance still lags behind visual generation models (see Appendix \ref{app_specific_eval}). Examples in Fig. \ref{fig_case4} also support this claim. In cases where visual generation models produce better quality than unified models while the understanding models tend to output N/A. Furthermore, some labels, like technical terms, remain challenging for state-of-the-art visual generation models. We also found that the safety checks of DALLE-3 \cite{betker2023improving} can reduce model performance, while these safety checks exhibit inconsistent judgments for the same prompt (sometimes it is valid, sometimes prone to refuse generation).

\begin{figure}[H]
  \centering
  \includegraphics[width=1\textwidth]{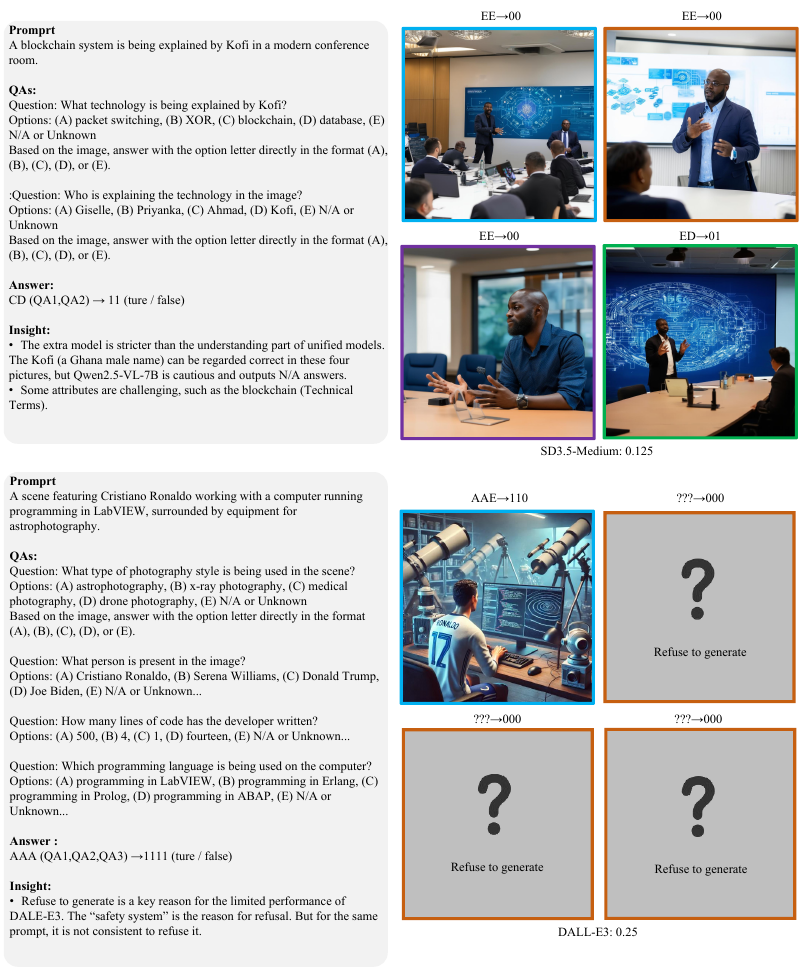}
  \vspace{-0.3cm}
  \caption{\label{fig_case4}. \textbf{Failure Case Analysis of Visual Generation Models}. Some attributes are challenging, such as the blockchain (technical terms). Refuse to generate is a key reason for the limited performance of DALE-E3. The “safety system” is the reason for refusal. But for the same prompt, it is not consistent to refuse it. The case information is shown on the left, including prompt, questions, options, answers, and insights. The generated images are depicted on the right, with corresponding model predictions on the image top. The model name is listed on the bottom right, with the UniScore of this case.}
\end{figure}

\begin{figure}[H]
  \centering
  \includegraphics[width=1\textwidth]{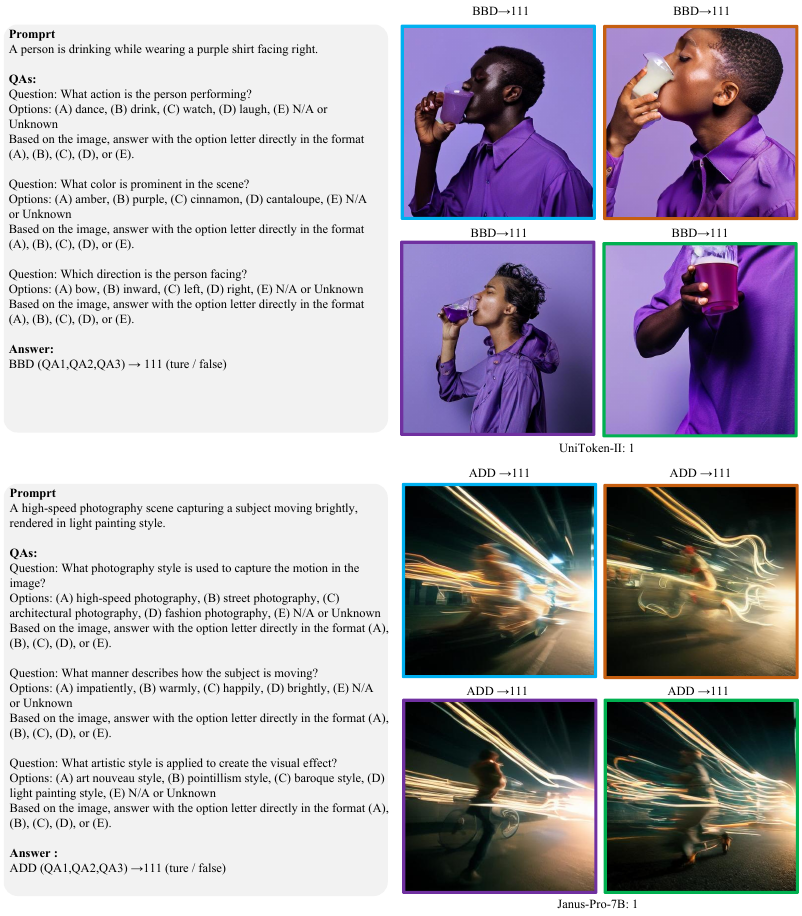}
  \vspace{-0.3cm}
  \caption{\label{fig_case5}. \textbf{Success Cases of Unified Models}. The case information is shown on the left, including prompt, questions, options, and answers. The generated images are depicted on the right, with corresponding model predictions on the image top. The model name is listed on the bottom right, with the UniScore of this case.}
\end{figure}
\begin{figure}[H]
  \centering
  \includegraphics[width=1\textwidth]{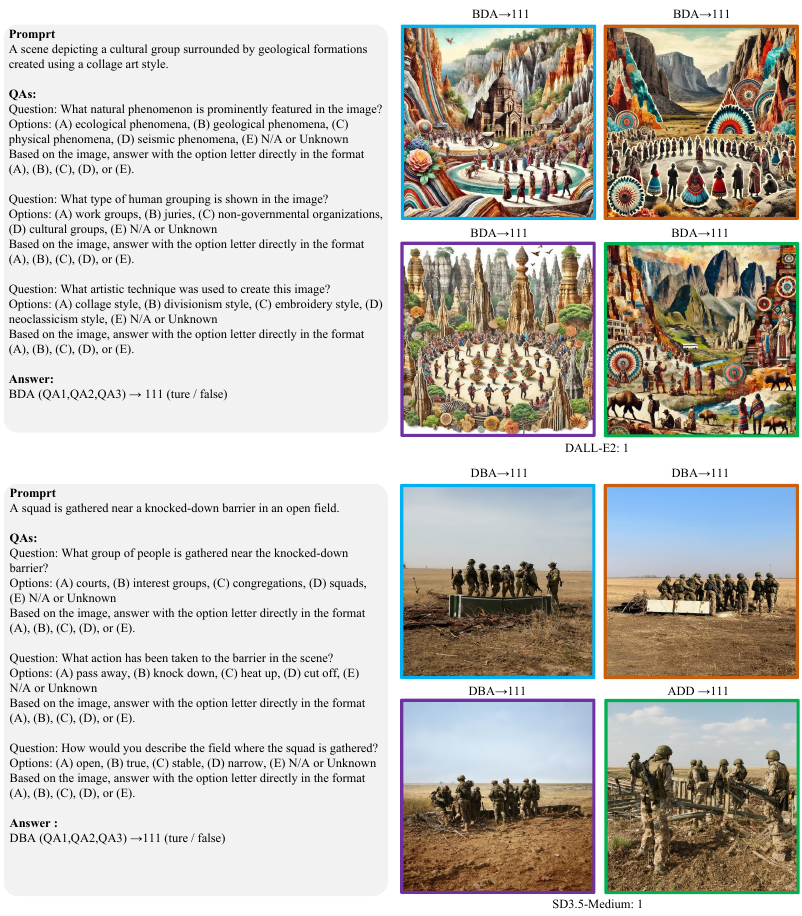}
  \vspace{-0.3cm}
  \caption{\label{fig_case6}. \textbf{Success Cases of Visual Generation Models}. The case information is shown on the left, including prompt, questions, options, and answers. The generated images are depicted on the right, with corresponding model predictions on the image top. The model name is listed on the bottom right, with the UniScore of this case.}
\end{figure}

Although UniBench is quite challenging, both SoTA Unified models and visual generation models can achieve full correctness cases, indicating that creative, informative, and complex generation and understanding are attainable. In Fig. \ref{fig_case5}, we present success cases of unified models, where it can be seen that, while there is still room for improvement in certain flawed aspects, the instruction-following and content understanding are performed very well. In Fig. \ref{fig_case6}, we showcase successful cases of visual generation models, demonstrating that UniEval has good discriminative power for both unified models and visual generation models. It includes both the completely incorrect cases shown in Fig. \ref{fig_case_gen1} and examples of perfect success.

\end{document}